\documentclass[sigconf, nonacm]{acmart}

\renewcommand\footnotetextcopyrightpermission[1]{}
\setcopyright{rightsretained}

\usepackage{bm}
\usepackage{subcaption}
\usepackage{nicefrac}
\usepackage{microtype}

\acmConference[]{}{}{}

\makeatletter
\def\@copyrightspace{\relax}
\makeatother

\begin{document}


\title[The Effectiveness of Discretization in Forecasting]{The Effectiveness of Discretization in Forecasting: An Empirical Study on Neural Time Series Models}

\author{Stephan Rabanser}
\affiliation{%
  \institution{AWS AI Labs}
}
\email{rabans@amazon.com}

\author{Tim Januschowski}
\affiliation{%
  \institution{AWS AI Labs}
}
\email{tjnsch@amazon.com}

\author{Valentin Flunkert}
\affiliation{%
  \institution{AWS AI Labs}
}
\email{flunkert@amazon.com}

\author{David Salinas}
\affiliation{%
  \institution{NAVER LABS Europe}
}
\email{david.salinas@naverlabs.com}

\author{Jan Gasthaus}
\affiliation{%
  \institution{AWS AI Labs}
}
\email{gasthaus@amazon.com}


\begin{abstract}
\noindent
Time series modeling techniques based on deep learning have seen many advancements in recent years, especially in data-abundant settings and with the central aim of learning global models that can extract patterns across multiple time series. While the crucial importance of appropriate data pre-processing and scaling has often been noted in prior work, most studies focus on improving model architectures. In this paper we empirically investigate the effect of data input and output transformations on the predictive performance of several neural forecasting architectures. In particular, we investigate the effectiveness of several forms of data \emph{binning}, i.e.\ converting real-valued time series into categorical ones, when combined with feed-forward, recurrent neural networks, and convolution-based sequence models. In many non-forecasting applications where these models have been very successful, the model inputs and outputs are categorical (e.g.\ words from a fixed vocabulary in natural language processing applications or quantized pixel color intensities in computer vision). For forecasting applications, where the time series are typically real-valued, various ad-hoc data transformations have been proposed, but have not been systematically compared. To remedy this, we evaluate the forecasting accuracy of instances of the aforementioned model classes when combined with different types of data scaling and binning. We find that binning almost always improves performance (compared to using normalized real-valued inputs), but that the particular type of binning chosen is of lesser importance.
\end{abstract}

%

\keywords{}

\maketitle

\section{Introduction}
\label{sec:intro}
Forecasting is the task to extrapolate time series into the future and it's a generally well-studied area (see ~\cite{hyndman2018forecasting} for an introduction). 
Forecasting has many important industrial applications in domains ranging from energy load~\cite{GEFCom14}, to e-commerce~\cite{bose2017probabilistic}, tourism~\cite{tourism11} and traffic~\cite{laptev2017}. Modern industrial applications often exhibit large panels of related time series, all of which need to 
be forecasted~\cite{janusch18}. The potential of ``neural'' forecasting models (i.e.\ models based on neural networks) in such contexts has long been exploited in applied industrial research, e.g.,~\cite{laptev2017,gasthaus2019probabilistic,flunkert2019deepar,wen2017}. Together with the overwhelming success of neural forecasting methods in the recent M4 competition~\cite{smyl2018m4}, this has convinced also formerly skeptical academics~\cite{Makridakis18,makridakis2018m4}.

Neural forecasting methods have seen many advancements in recent years, especially in data-abundant settings and with the central aim of learning a single \emph{global} model 
over the entire panel of time series to then extract patterns across multiple time series. Note that global models are distinct from multi-variate models. Global models produce univariate forecasts, i.e., forecast each time series member of the panel of time series independently, but parameters of the model are estimated over the entire panels. Multi-variate time series models explicitly estimate the dependence structure of the time series in the panel. While multi-variate models form an important area and initial work exists for neural forecasting methods (e.g.,~\cite{salinas2019}), much further work 
is needed and we focus on univariate time series forecasting in this article.

Many deep learning architectures that have seen success in other domains (e.g.\ computer vision or natural language processing) have been adapted to and evaluated in the forecasting setting, ranging from simple feed forward models, convolutional neural networks (CNNs), in particular using 1-dimensional dilated causal convolutions \citep{oord2016wavenet, borovykh2017, wen2017, bai2018}, recurrent neural networks (RNNs) \citep{flunkert2019deepar, mukherjee2018,smyl2018m4}, and attention-based models \citep{li2019, lim2020, vaswani2017attention}.

While some prior work has only considered the point forecasting setting, we focus on models that can produce probabilistic forecasts, i.e.\ forecasts that quantify the uncertainty over future events by estimating a probability distribution over future trajectories. Such probabilistic forecasts can be used for decision making under uncertainty, which is typically the ultimate goal in practical applications. To that end, the various aforementioned deep learning architectures have been combined with techniques for modeling probabilistic outputs. These techniques range from parametric distributions and parametric mixtures \citep{flunkert2019deepar, mukherjee2018}, over quantile regression-based techniques like quantile grids \citep{wen2017}, to parametric quantile functions models \citep{gasthaus2019probabilistic}, semi-parametric probability integral transform / copula based techniques \citep{salinas2019, wen2019}, and approaches based on discretization/bucketing \citep{oord2016wavenet}.

Recent developments in neural time series forecasting have mostly focused on improving model architectures \citep{li2019, lim2020, nbeats, LSTNet, sen2019think}, and developing strategies for modeling the probabilistic outputs in these models \citep{gasthaus2019probabilistic, salinas2019, wen2019} (see \citet{faloutsos19forecasting2} for a recent overview). 
The work on global neural models for time series forecasting \citep{flunkert2019deepar} has often hinted at (but not explored in detail) the importance of careful data pre-processing for learning across time series, especially when modeling datasets with heterogeneous magnitudes (e.g. in the retail demand forecasting setting). While different strategies for handling the differences in magnitudes across time series have been proposed, including mean/median scaling, standardization, bijective transformations (e.g. log or Box-Cox transforms), and discretization, a thorough and systematic empirical evaluation of the impact on predictive performance and training stability that input and output representations have relative to the core forecasting model has not been performed. 
The study presented in this paper shines some light on this question by performing an empirical comparison of multiple different input and output transformation techniques---with a particular focus on discretizing transformations---when combined with commonly-used neural forecasting architectures. It complements empirical studies evaluating the impact of other architectural choices, e.g.\ the extensive study of RNN models for forecasting conducted by \citet{hewamalage2019recurrent}. 

The main finding and core contribution of our empirical study is the importance of discretization of inputs and outputs as a general technique for neural forecasting models. Our 
experimental results show that binning techniques 
improve performance of forecasting models almost independently of the architecture of the neural network. This is mildly surprising since the inputs and target time series in the forecasting setting are real values, and thus endowed with a natural total order and a notation of distance, which the model does not have access to after discretization.
Further, typical forecasting accuracy measures~\cite{gneiting2007probabilistic,hyndman2006accuracy} also rely on notions of distance, giving higher scores to forecasts that are ``close'' to the true values. That giving up this order through discretization and adopting a loss function in 
neural network models that does not take this order into account explicitly leads to superior accuracy, is curious.

The rest of the paper is structured as follows: In Sec. \ref{sec:preliminaries} we first describe the general forecasting task and setup we consider; Sec. \ref{sec:method} describes the different data transformations and models that we compare; Sec. \ref{sec:exp} contains the experimental results, which we discuss further in Sec. \ref{sec:diss}. 
We discuss related work in Sec. \ref{sec:related} and present some conclusions in Sec. \ref{sec:con}.

\section{Preliminaries}
\label{sec:preliminaries}
Our study explores the following commonly-used setup for forecasting problems. We are given a set $Z = \{z_{i,1:T_i}\}_{i=1}^{N}$ of $N$ univariate time series. Each time series $z_{i,1:T_i} = (z_{i,1}, z_{i,2}, \ldots, z_{i,T_i})$ is composed of $T_i$ consecutive values $z_{i,t} \in \mathbb{R}$ which are assumed to be equally spaced.
In addition to the \emph{target} time series $z_{i,t}$ (i.e.\ the ones we are trying to predict the future of), the methods we consider can optionally make use of a set of associated covariates $X = \{\bm{x}_{i,1:T_{i}+\tau}\}$ with $\bm{x}_{i, t} \in \mathbb{R}^D$, which are required to be available until time point $T_{i}+ \tau$ with $\tau$ being the prediction horizon of the forecast. Note that in this paper we will exclusively focus on applying transformations to the target time series values $z_{i,t}$, considering the covariates $\bm{x}_{i,t}$ given and fixed. In practice, the covariates are often synthetically constructed (e.g.\ date-dependent dummy variables) and require no further processing, or similar normalization techniques as we discuss for the target time series can be applied.

Our goal is to model the joint conditional probability distribution over $z_{i, T_i+1:T_i + \tau}$ for each time series $i$, given its past values $z_{i,1:T_i}$ and the observed additional covariates $\bm{x}_{i,1:T_{i}+\tau}$.
\emph{Global} neural forecasting models achieve this by parametrizing this conditional distribution using a neural network $\mathcal{M}_\Theta$, whose parameters $\Theta$ are learned \emph{jointly} from the entire data set $(Z, X)$.
In particular, for each time series $i$ we have,
\begin{equation}
	p(z_{i,T_{i}+1:T_{i}+\tau}\ |\ z_{i,1:T_i}, \bm{x}_{i,1:T_{i}+\tau}) = \mathcal{M}_\Theta(z_{i,1:T_i}, \bm{x}_{i,1:T_{i}+\tau}),
	\label{eq:prob_predict}
\end{equation}
and the parameters $\Theta$ are learned by optimizing some scoring rule $\mathcal{L}$ (often negative log-likelihood) measuring the compatibility of the model with the observed data over the training data set, i.e.\ $\Theta^\star = \text{argmin}_\Theta\, \sum_i \mathcal{L}(\mathcal{M}_\Theta, (z_{i,1:T_i}, \bm{x}_{i, 1:T_i}))$.

Note that in Eq. \eqref{eq:prob_predict} the values of the target time series $z_i$ appear in both the conditioning set and the predicted variables. We refer to a transformation that is applied to the variables in the conditioning set as an \emph{input transformation}, and to a transformation that affects the predicted distribution as an \emph{output transformation}. The resulting transformed values are \emph{input} and \emph{output representations}, respectively.
To illustrate the idea, consider a simple Markov model that predicts the next value $z_{t+1}$ conditioned on the preceding value $z_t$. Instead of directly modeling $p_z(z_{t+1}|z_t)$ we can compute transformed inputs $x_t = f_\phi(z_t)$ and outputs $y_{t+1} = g_\psi(z_{t+1})$ and model $p_y(y_{t+1}|x_t)$ instead. If $g_\psi$ is invertible we can generate samples from the predictive distribution by sampling $y_{t+1} \sim p_y(\cdot|x_t)$ and computing $z_{t+1} = g^{-1}_\psi(y_{t+1})$. If necessary (e.g.\ for computing the training loss) we can also evaluate the density, 
\[
p_z(z_{t+1}|z_t) = p_y(g_\psi(z_{t+1})|f_\phi(z_t))\left|\frac{\mathrm{d} g_\psi(z_{t+1})}{\mathrm{d} z_{t+1}}\right|.
\]
Note that while using the same transformation for both input and output, i.e.\ $f_\phi = g_\psi$, is commonly done (e.g.\ by pre-processing the data before applying the model), this is not necessary. In particular, the input transformation $f_\phi$ is not required to be invertible in general.

When global models are used in the series forecasting setting, the parameters $\Theta$ are learned jointly over the data set, but the parameters $\phi$ and $\psi$ of the input and output transformations are typically allowed to vary per time series, e.g.\ by estimating them on data preceding the training range. For example, the mean-scaling approach employed by the DeepAR method~\cite{flunkert2019deepar} estimates $\phi_i = m_i = \frac{1}{T_i}\sum_{t=1}^{T_i} |z_{i, t}|$ and then sets $f_{\phi_i}(z) = \frac{z}{m_i}$. DeepAR uses no output transformation, but includes $m_i$ as an additional input to the model.

\section{Methods}
\label{sec:method}
Next, we introduce the main objects in our study, the input and output transformations, and the neural network models. 

\subsection{Transformations}
The transformations of data that we apply in our empirical study range for schemes to rescale the time series in the panel to 
discretization and other transformations. We describe these next in detail.
\subsubsection{Scaling}
\label{sec:scaling}
When training global forecasting models on datasets with heterogeneous scales, accounting for the difference in scales between time series in some way is of critical importance for obtaining good predictive performance. 
Firstly, it is desirable for the models to learn scale-invariant patterns, especially seasonal behavior (e.g.\ the seasonal sales pattern for a text book is largely independent of how popular the specific book is). Secondly, neural network models with saturating non-linearities are very sensitive to the scale of their inputs, leading to slow convergence (or convergence to undesirable optima) if the scale of their inputs is not carefully controlled. 


A common approach for addressing the challenge of heterogeneous scales is to apply an affine transformation to each time series, i.e.\ $z_{i,t}' = (z_{i, t} - b_i)/a_i$, where the parameters for the transformation are chosen for each time series independently. 
Here, as a representative member of this family of transformations, we use the mean scaling ($\texttt{ms}$) scheme employed e.g.\ by DeepAR \citep{flunkert2019deepar}, which seems to be effective in many practical settings. In particular, we set $a_i = \frac{1}{T_i} \sum_{t=1}^{T_i} |z_{i,t}|$ and $b_i =0$. 
Other common choices include min-max scaling ($a_i = \max(z_{i,1:T_i}) - \min(z_{i,1:T_i})$, $b=\min(z_{i,1:T_i}))$ and standardization ($a_i = \text{stddev}(z_{i, 1:T_i})$, $b_i = \text{mean}(z_{i,1:T_i})$). 
In practice, variants that use a more robust estimate of the scale, e.g.\ the trimmed mean or the median have also been employed.

\subsubsection{Continuous Transformations}
In addition to the affine transformations described above, other continuous transformations, such as power transforms, e.g. the Box-Cox transform $z'_{i,t} = (z_{i,t}^{\lambda_i} - 1)/\lambda_i$ \citep{Box.Cox1964} or the log transform (which is its limit as $\lambda \rightarrow 0$) are commonly applied not only in combination with ``classical'' forecasting techniques (e.g.\ ARIMA, exponential smoothing, or linear regression models, see~\cite{hyndman2018forecasting} for an introduction), but also in combination with deep learning techniques.
The Box-Cox transform was originally introduced as a ``Gaussianizing'' transform, i.e.\ a transformation that makes the data distribution ``more Gaussian'', so that techniques that assume Gaussian noise can more readily be applied. Here we will consider an alternative technique for transforming the marginal distributions into a form more amenable to modeling, based on the probability integral transform, similar to the approach proposed by \citep{salinas2019}.

The \emph{probability integral transform} (PIT) is the transformation that maps a random variable $X$ through its cumulative distribution function, i.e.\ $Y = F_X(X)$, resulting in a transformed variable $Y$ with uniform distribution.
In the setting we are considering here, an (approximate) probability integral transform (\texttt{pit}) can be used to make the empirical marginal distribution of values in each time series (approximately) uniform. In particular, we apply the transform $z_{i,t}' = \hat{F}_i(z_{i,t})$, where $\hat{F}_i$ is the empirical cumulative distribution function estimated from $z_{i, 1:T_i}$.
The PIT is effectively the non-discretizing (i.e.\ producing real-valued instead of categorical outputs) analogue of the quantile binning transform discussed below. In order to make these two approaches directly comparable, we combine the PIT input transform with an additional two-layer input transformation network, which performs a function similar to the embedding layer used in conjunction with the discrete inputs. In principle, such a network could learn to implement a binning operation in the first layer, while the second layer performs the embedding, making this setup at least as expressive as quantile binning followed by an embedding layer.

\subsubsection{Discretizing Transformations}

Binning is a form of data discretization (also called \emph{quantization}) into a set of buckets with disjoint support, that is widely used in machine learning as a feature engineering technique. Formally, we define a function $b: \mathbb{R} \rightarrow \{1,2,\ldots,B\}$ which maps a real-valued input to a discrete output with $B \in \mathbb{N}$ distinct bin values. Each of the possible output values $b \in \{1, \ldots,B\}$ is tied to a specific interval (``bucket'') $S_b = [l_{b-1}, l_{b})$ into which real-valued inputs can fall, with edge cases $l_0 = -\infty$ and $l_B = \infty$. The quantization transform $b$ then maps a real-valued input to its bucket index, i.e. $b(x) = b$ iff $x \in S_b$. In case the input domain happens to be a subset of $\mathbb{R}$ we can adjust the edge cases accordingly. 

In order to also define a reconstruction function $s: \{1,2,\ldots,B\} \rightarrow \mathbb{R}$ which transforms a discrete bucket value back to the original real-valued domain, we associate each bucket $b \in {1, \ldots, B}$ with a value \emph{reconstruction value} $c_b \in \mathbb{R}$ and set $s(b) = c_b$. Given such reconstruction values $c_1 \leq c_2 \leq \ldots \leq c_B$ and assuming squared error as the loss function, it can be shown that the reconstruction error is minimized by choosing the bin edges as $l_b = (c_{b} + c_{b+1})/2$. Note that optimal reconstruction values $c_b$, in the sense of minimizing squared reconstruction error, can also be obtained (given fixed bin edges $l_b$) by setting $c_b = \int_{l_{b-1}}^{l_b} x f(x)dx /\int_{l_{b-1}}^{l_b} f(x)$ where $f(x)$ is the density of the real-valued inputs. Iterating the steps for optimizing the reconstruction values and bin edges is the Lloyd-Max algorithm for obtaining optimal quantizers. 


Different strategies have been developed for selecting appropriate bin edges (or reconstruction values $c_b$ and choosing the edges as midpoints as described above) and we will consider two strategies as part of this paper: equally-spaced binning and quantile binning. In equally-spaced (linear) binning, a part of the input $I = [x_{\text{min}}, x_{\text{max}}] \subset \mathbb{R}$ is divided into $B-2$ intervals with equal width, i.e. 
for $b \in \{1, \ldots, B-1\}$, $l_b = x_\text{min} + (b-1)(x_\text{max} - x_\text{min})/(B-2)$. 
In contrast to linear binning, quantile binning makes use of the underlying cumulative distribution function (CDF) $F_Z$ to construct a binning such that the number of data points falling into each bin is (approximately) equal. 
In quantile binning, we first create a list of equally spaced quantiles $q = (q_1, q_2, \ldots, q_B)$ where $q_j - q_{j+1} = \frac{1}{B}$ and $q_j \in [0,1]\ \forall j \in \{1,2,\ldots,B\}$. Then, we can obtain the quantile-based bin representations $c_b$ by evaluating the quantile function $F^{-1}_Z$ for each $q_b$, i.e.\ $c_b = F^{-1}_Z(q_b)\ \forall b \in \{1,2,\ldots,B\}$, ensuring that all buckets contain approximately the same number of samples.


A nowadays well-established strategy for using categorical inputs with deep learning models we adopt here is to use an embedding layer as the first layer in the model, which maps each categorical input $b \in \{1, 2, \ldots, B\}$ to a vector $e_b \in \mathbb{R}^E$ that is learned together with the weights of the network using gradient descent. 

As part of this study we consider two main binning strategies for time series forecasting: local absolute binning, where the bins are computed for each time series separately, and global relative binning, where the bins are computed jointly for the entire dataset, after scaling each time series individually.

\paragraph{Local Absolute Binning ($\texttt{lab}$)}

In local absolute binning, each time series is binned individually. This involves computing the the bin edges $l_i = (l_{i,1}, l_{i,2}, \ldots, l_{i,B})$ for each time series and then binning each time series $z'_{i,t} = b(z_{i,t}|l_i)$. Since each time series is binned using its own set of bin edges and each time series is mapped to the same set of bin identifiers $\{1,2,\ldots,B\}$, local absolute binning effectively acts as a scaling mechanism. 

\paragraph{Global Relative Binning ($\texttt{grb}$)}

\begin{figure*}[t]
    \centering
    \begin{subfigure}[t]{0.30\textwidth}
        \centering
        \includegraphics[width=.98\linewidth]{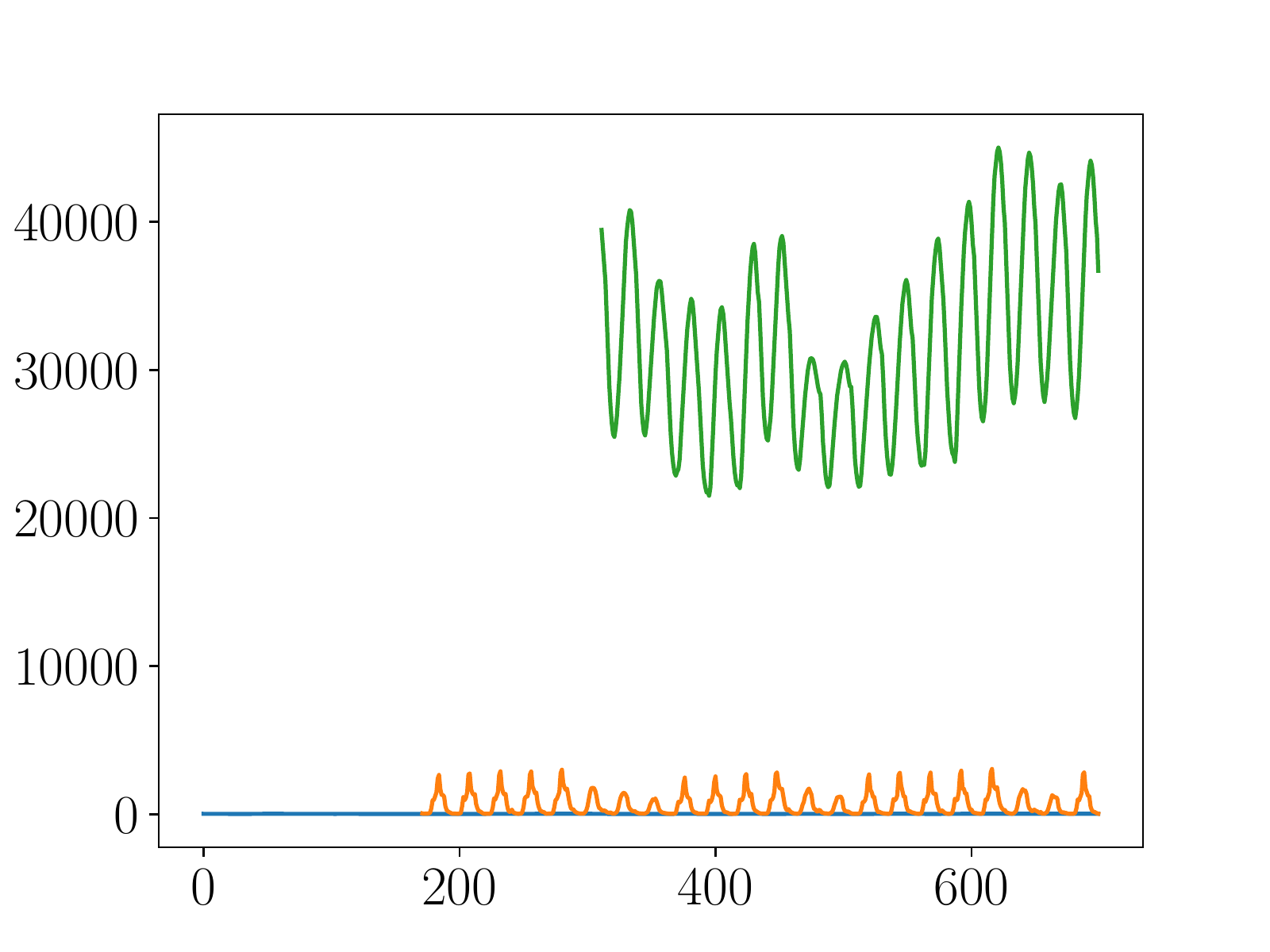}
        \caption{Original unprocessed time series.}
    \end{subfigure}%
    \hspace{10pt}
    \begin{subfigure}[t]{0.30\textwidth}
        \centering
        \includegraphics[width=.98\linewidth]{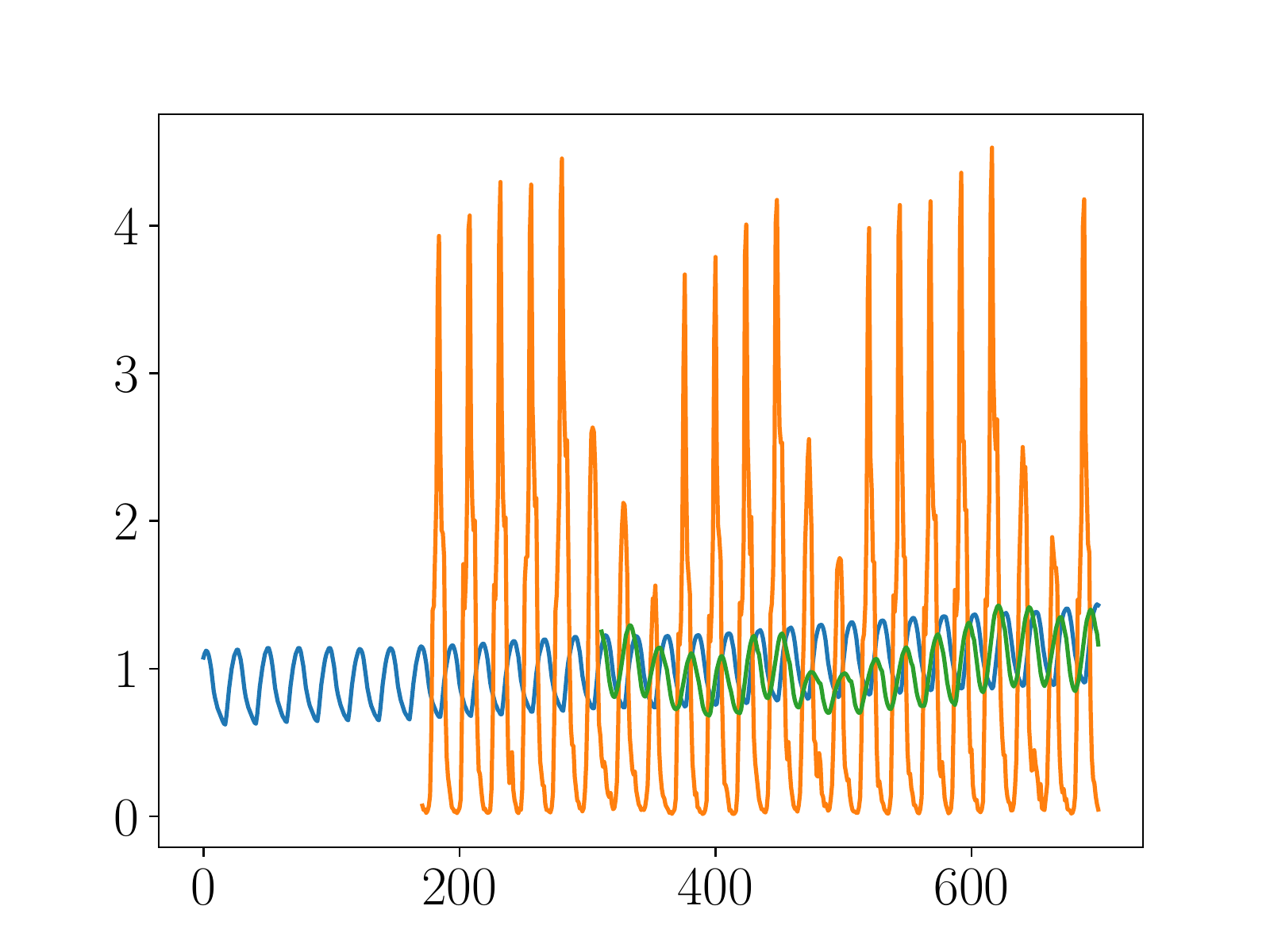}
        \caption{Time series after mean scaling.}
    \end{subfigure}
    \hspace{10pt}
    \begin{subfigure}[t]{0.30\textwidth}
        \centering
        \includegraphics[width=.98\linewidth]{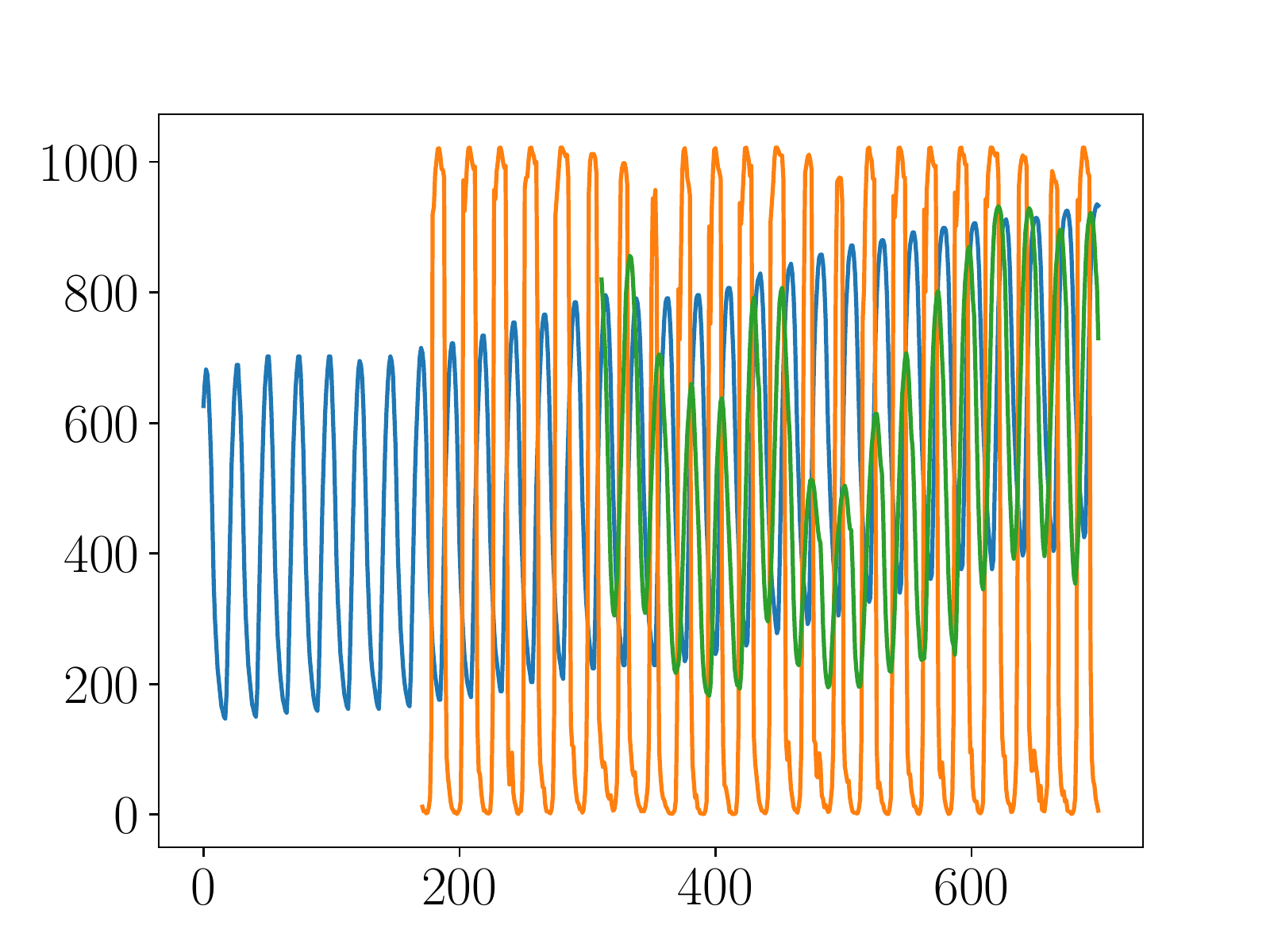}
        \caption{Time series after global binning.}
    \end{subfigure}
    
    \begin{subfigure}[t]{0.30\textwidth}
        \centering
        \includegraphics[width=.98\linewidth]{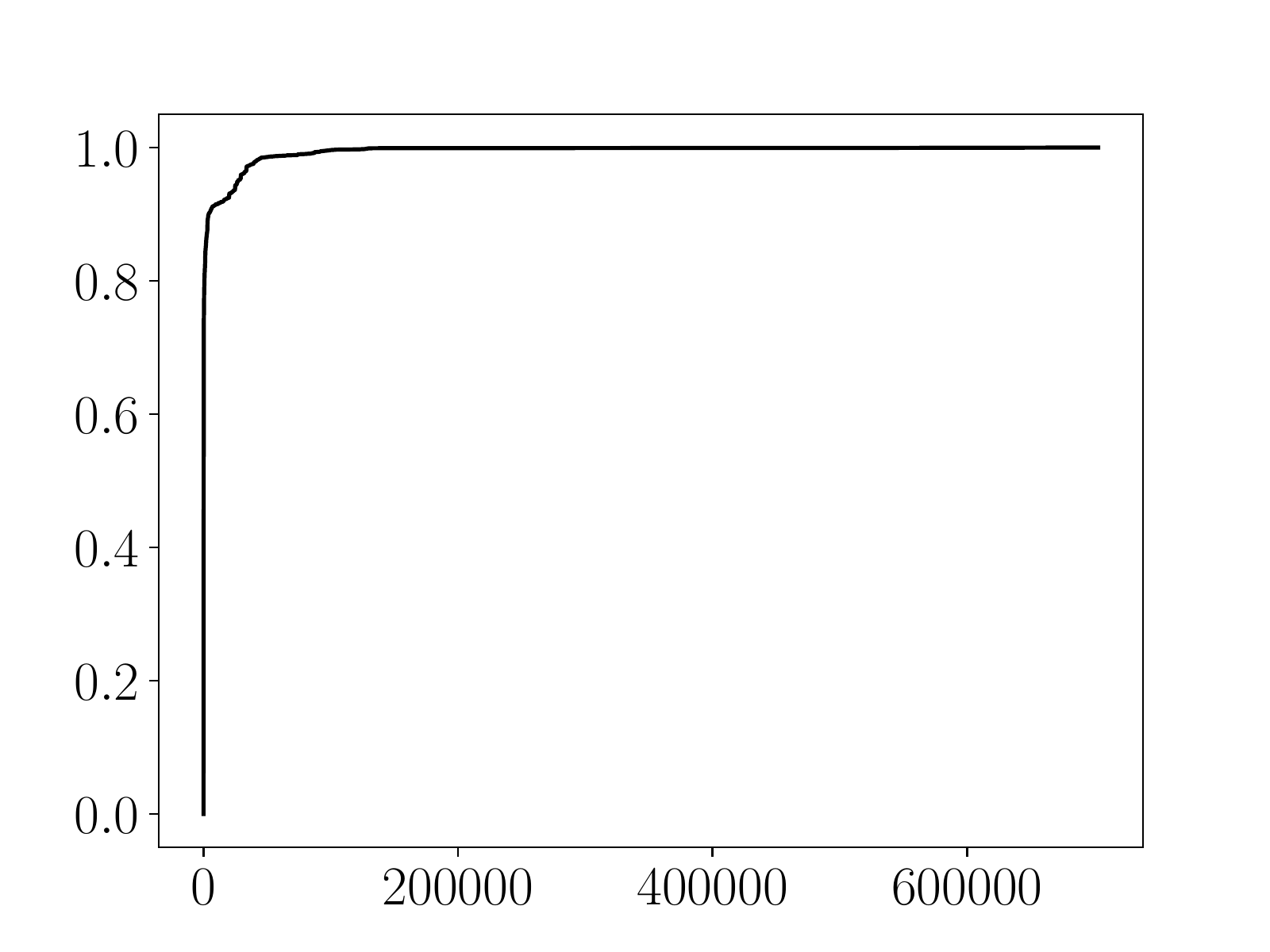}
        \caption{Original CDF. We can clearly see that the distribution contains only a few time series with very large scales.}
    \end{subfigure}%
    \hspace{10pt}
    \begin{subfigure}[t]{0.30\textwidth}
        \centering
        \includegraphics[width=.98\linewidth]{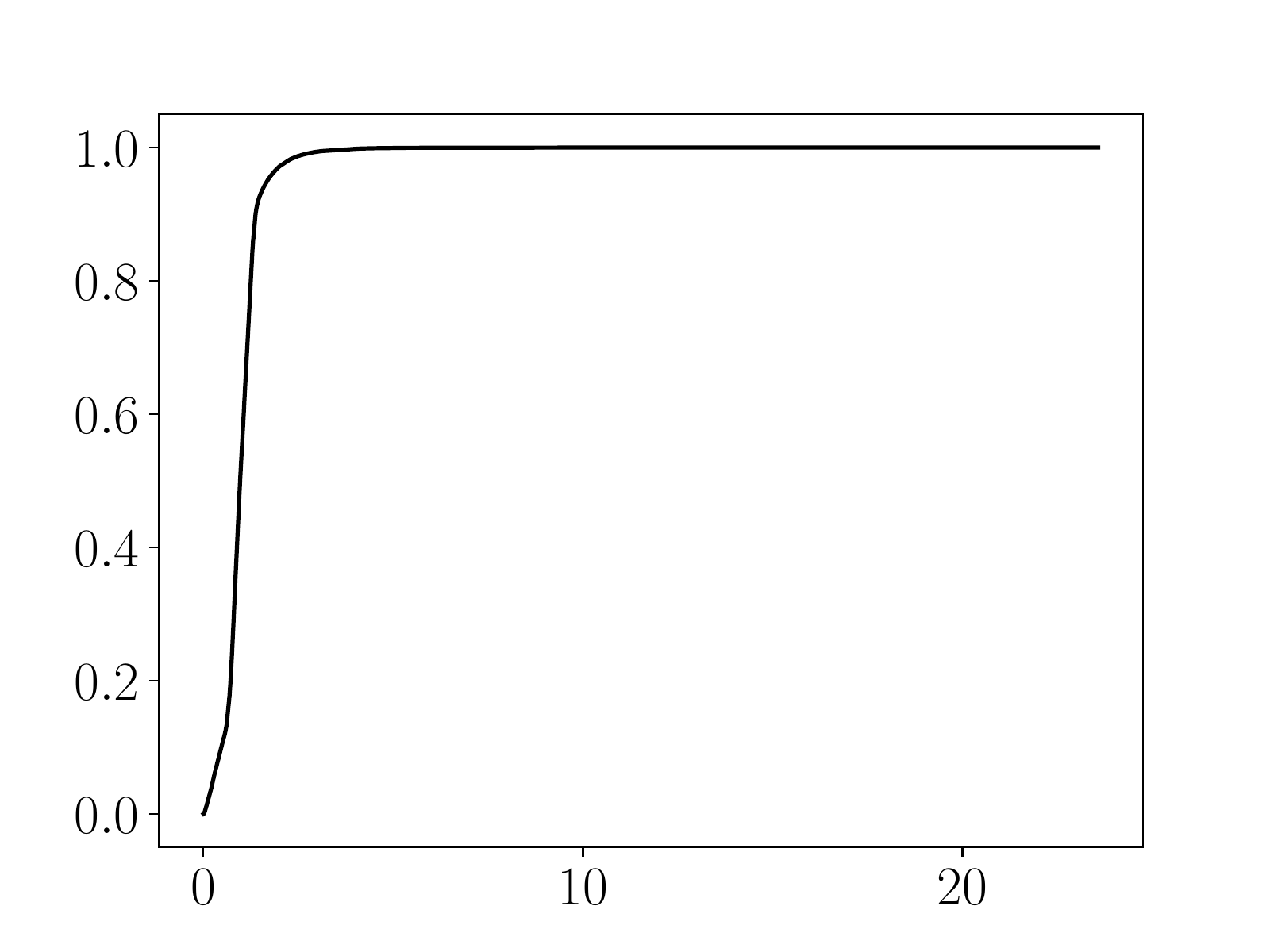}
        \caption{CDF after mean scaling. While mean scaling lessens the effect that large time series have on the CDF, a few outliers remain.}
    \end{subfigure}
    \hspace{10pt}
    \begin{subfigure}[t]{0.32\textwidth}
        \centering
        \includegraphics[width=.98\linewidth]{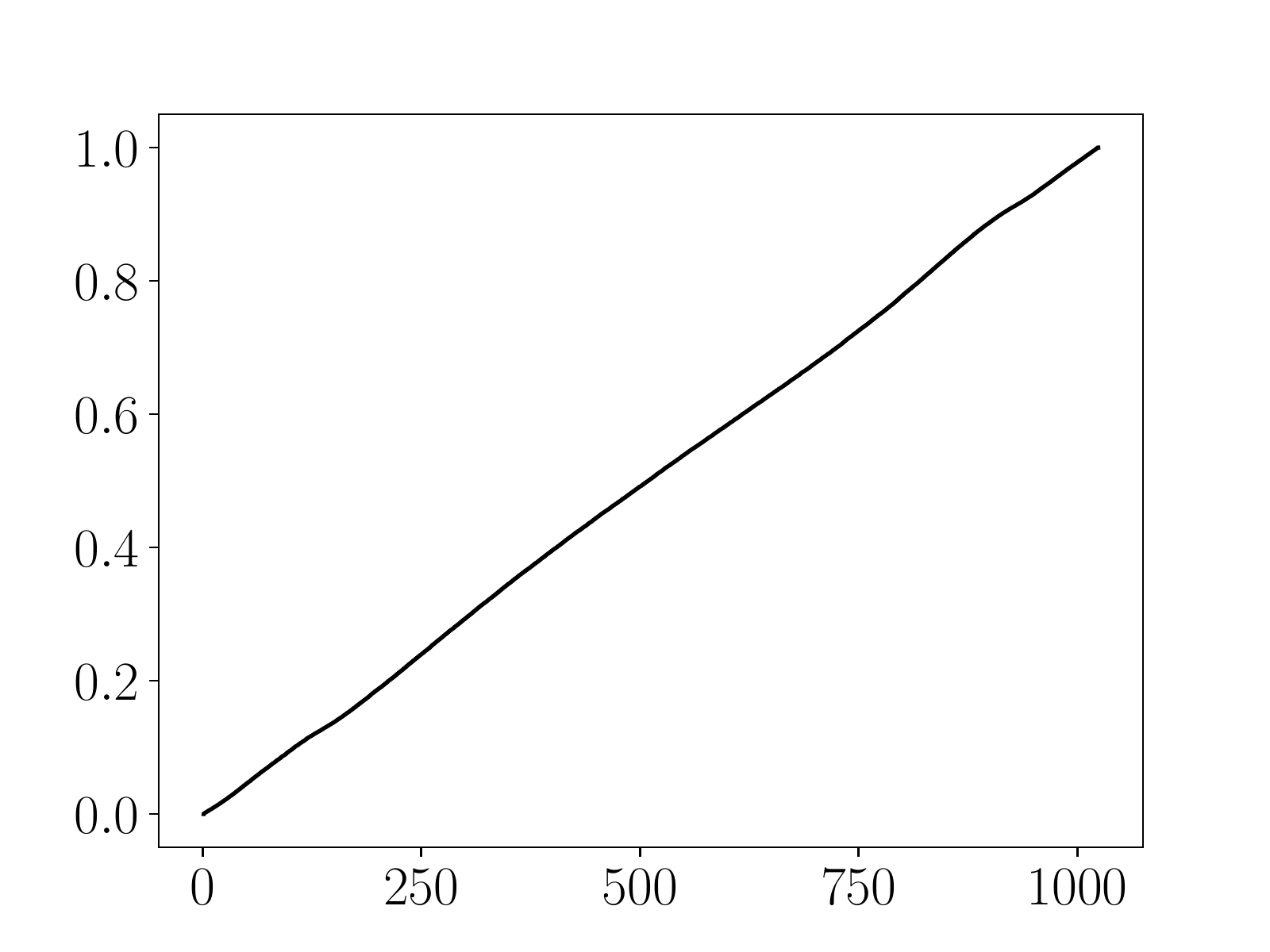}
        \caption{CDF after global binning. Since quantile binning uses the quantile function to transform the time series, the resulting CDF corresponds to a uniform distribution the bins.}
    \end{subfigure}
 
    \caption{Global relative binning example on \texttt{m4\_hourly} with 1024 bins and quantile binning. Plots (a)-(c) show how 3 randomly picked time series pass through the scaling and binning transformations and plots (d)-(f) show the respective CDF distributions over the entire training set.}
    \label{fig:scaling}
\end{figure*}

\begin{figure*}[t]
    \centering
    \begin{subfigure}[t]{0.30\textwidth}
        \centering
        \includegraphics[width=.98\linewidth]{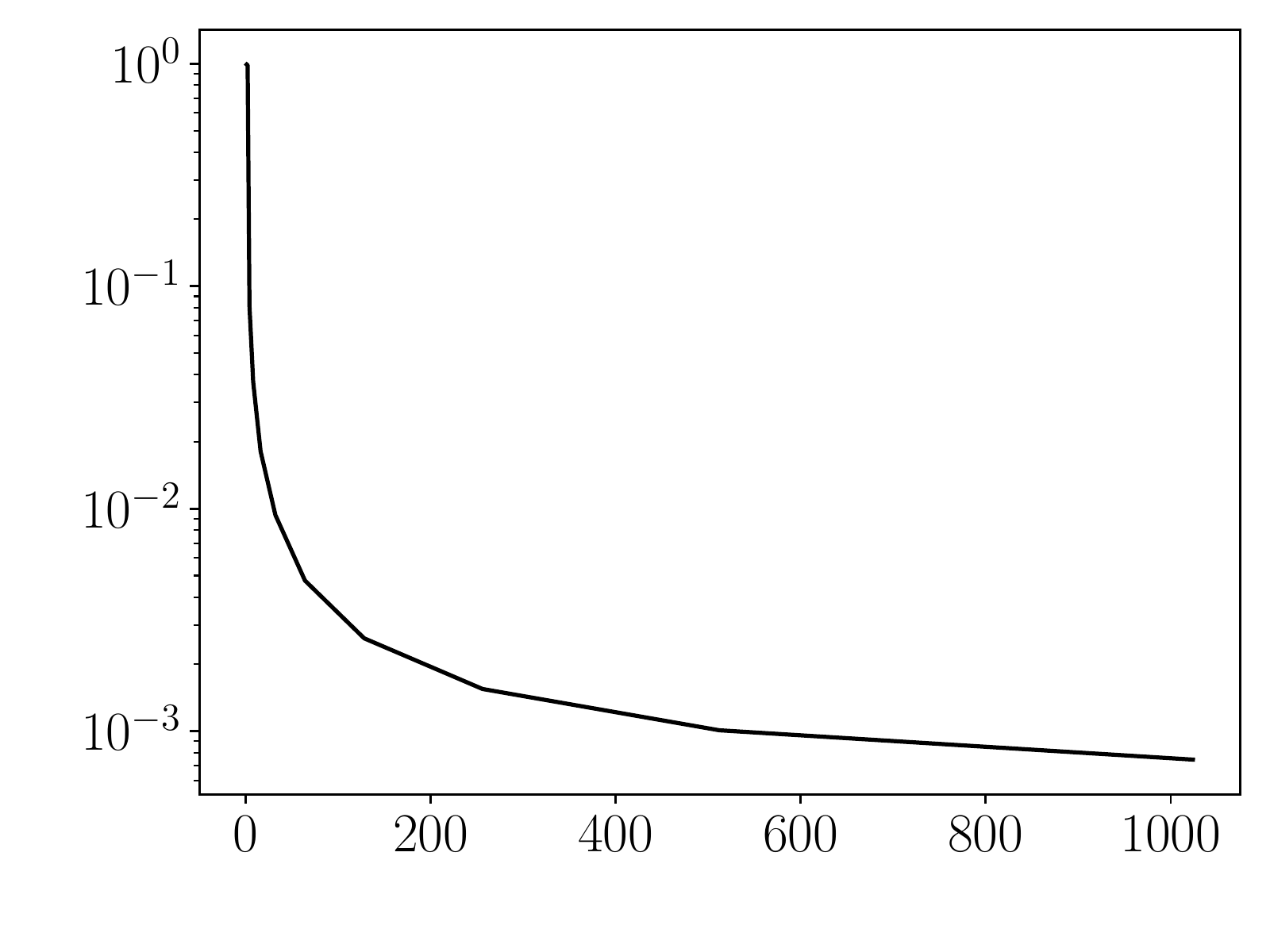}
        \caption{Relative reconstruction loss over all time series decreases with increasing number of bins.}
    \end{subfigure}%
    \hspace{10pt}
    \begin{subfigure}[t]{0.32\textwidth}
        \centering
        \includegraphics[width=.98\linewidth]{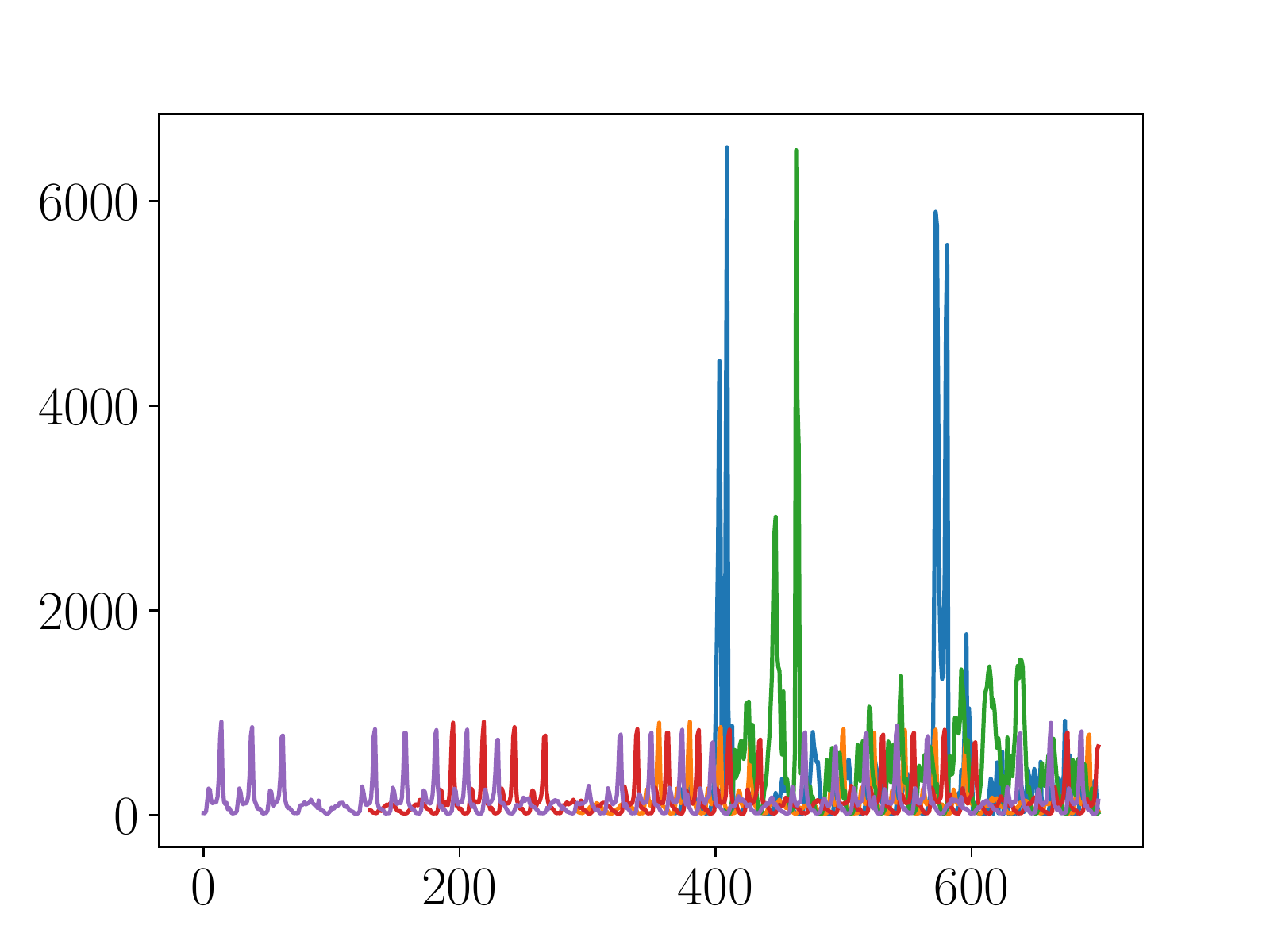}
        \caption{Top-5 time series with largest reconstruction loss. These series show major outliers.}
    \end{subfigure}
    \hspace{10pt}
    \begin{subfigure}[t]{0.30\textwidth}
        \centering
        \includegraphics[width=.98\linewidth]{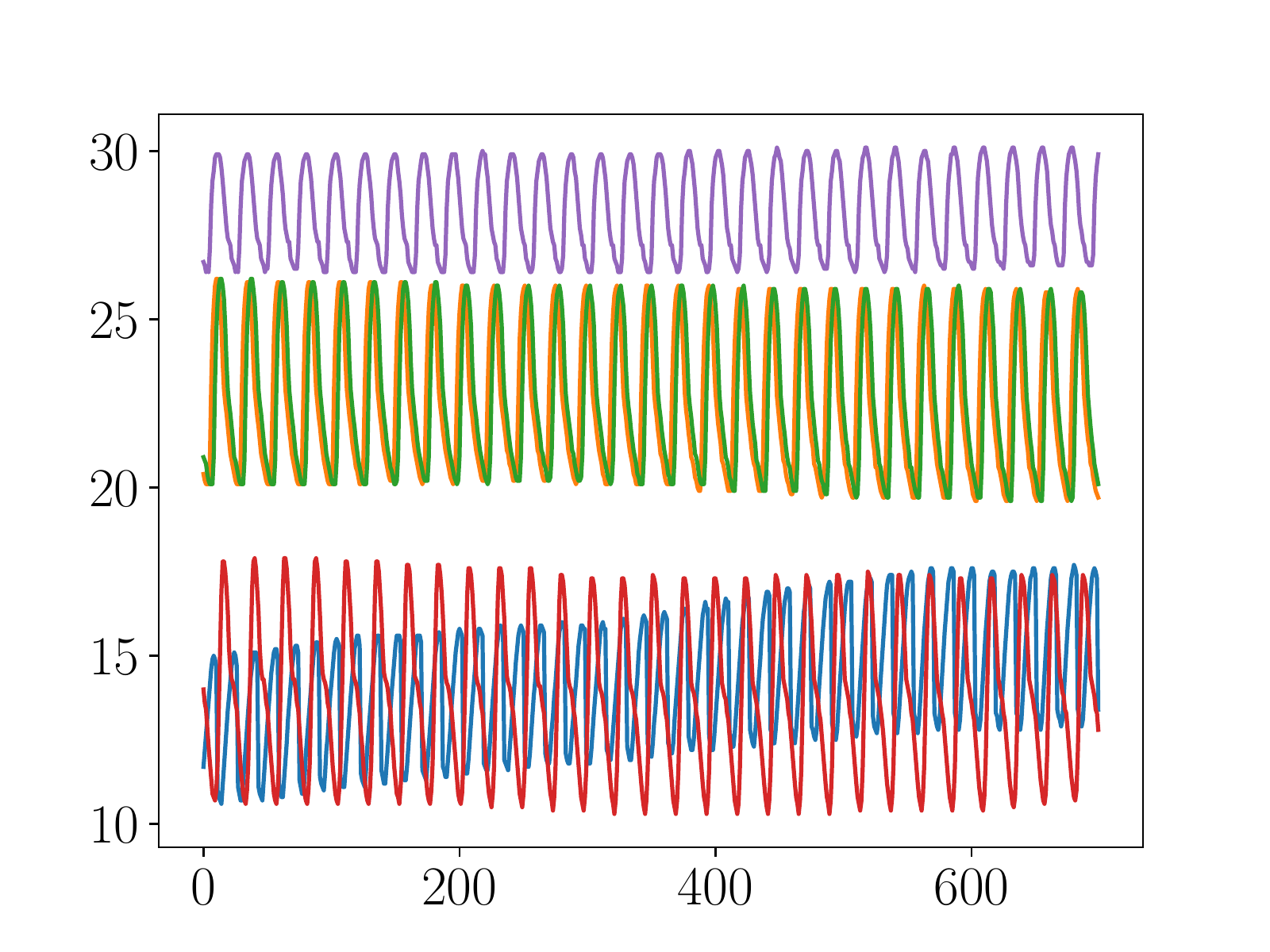}
        \caption{Top-5 time series with smallest reconstruction loss. These series are highly regular and periodic.}
    \end{subfigure}
    \vspace{10pt}
    \begin{subfigure}[t]{0.30\textwidth}
        \centering
        \includegraphics[width=.98\linewidth]{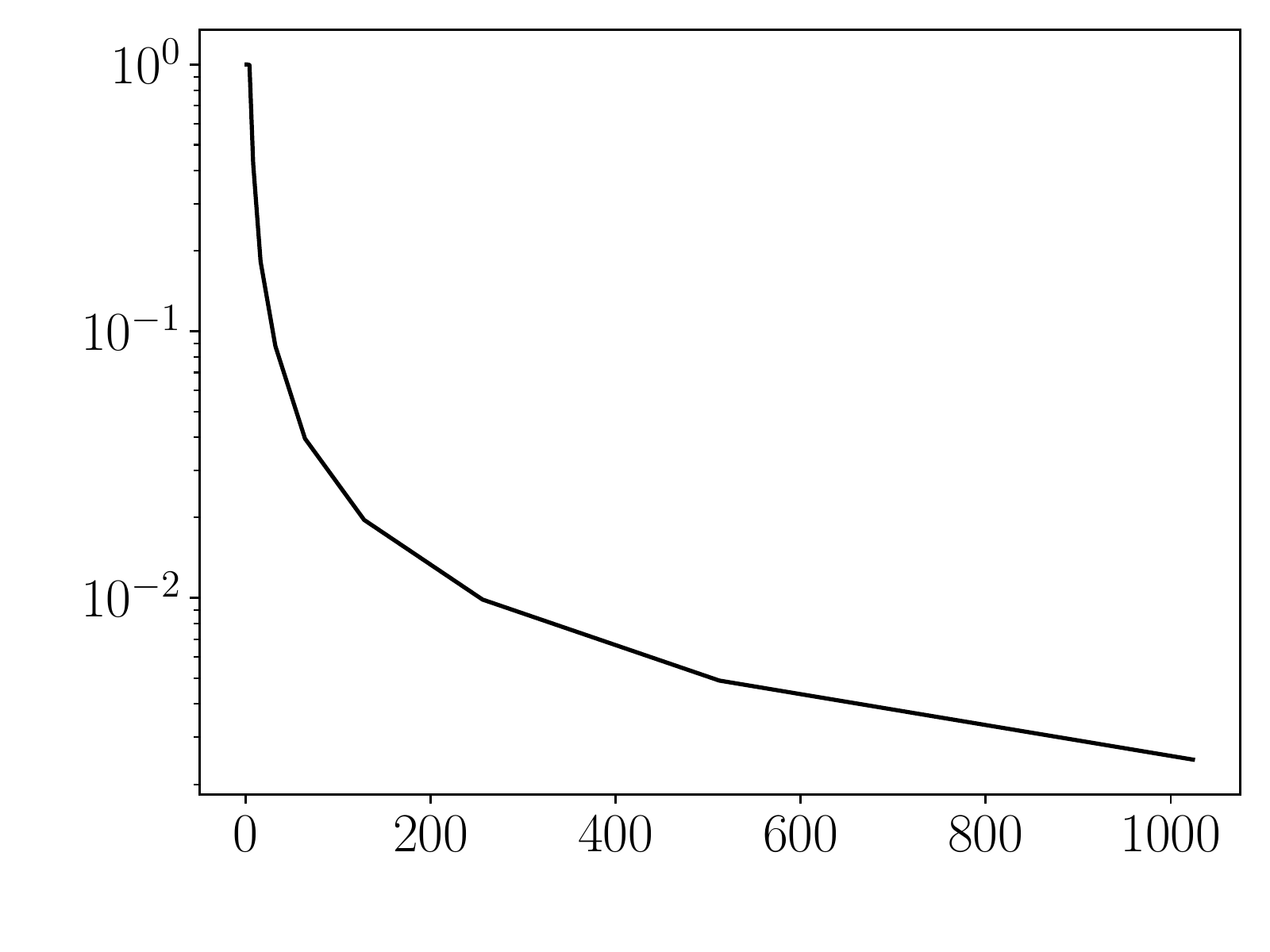}
        \caption{Relative reconstruction loss over all time series decreases with increasing number of bins.}
    \end{subfigure}%
    \hspace{10pt}
    \begin{subfigure}[t]{0.30\textwidth}
        \centering
        \includegraphics[width=.98\linewidth]{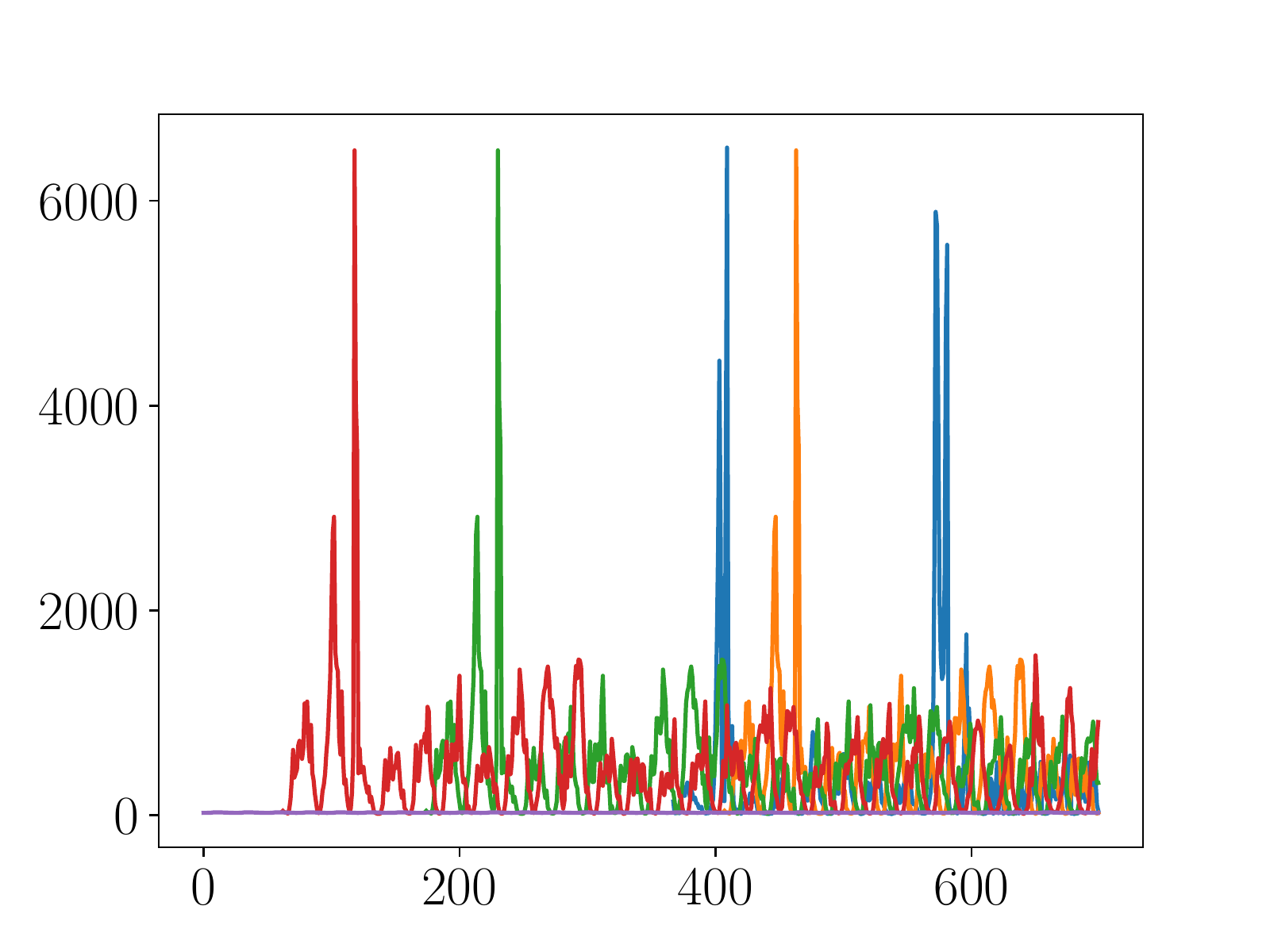}
        \caption{Top-5 time series with largest reconstruction loss. These series show major outliers.}
    \end{subfigure}
    \hspace{10pt}
    \begin{subfigure}[t]{0.30\textwidth}
        \centering
        \includegraphics[width=.98\linewidth]{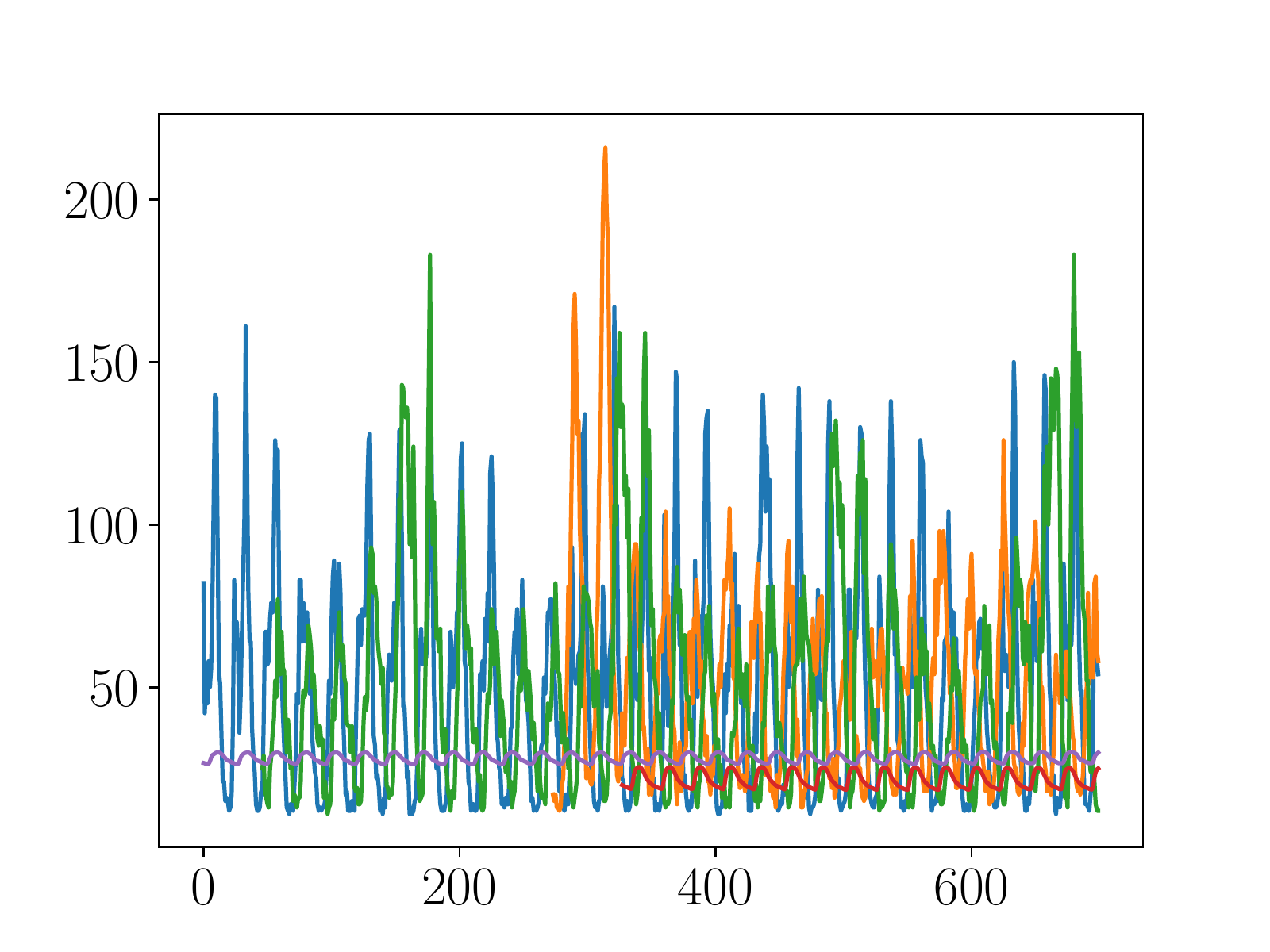}
        \caption{Top-5 time series with smallest reconstruction loss.}
    \end{subfigure}
 
    \caption{Time series reconstruction using global relative binning on \texttt{m4\_hourly} with varying bin sizes bins. Plots (a)-(c) show quantile binning results, plots (d)-(f) linear binning results. It is evident that quantile binning achieves a smaller reconstruction error than linear binning at a given number of bins.}
    \label{fig:reconstr}
\end{figure*}

In global relative binning, all time series are first rescaled and then binned with one global binning. In particular, we use the mean scaling approach described before to scale each time series.
We can then estimate one single set of bin edges $l = (l_{1}, l_{2}, \ldots, l_{B})$ over the entire collection of scaled time series and bin every time series according to these bin edges $z''_{i,t} = b(z'_{i,t}|l)$ where $z'_{i,t}$ corresponds to the scaled time series. A visual example of global relative binning is shown in Figure \ref{fig:scaling}, while an analysis on the reconstruction loss incurred via global binning is shown in Figure \ref{fig:reconstr}.

\paragraph{Hybrid Binning ($\texttt{hyb}([\cdots])$)}
In addition to pure local absolute or global relative binnings, one can compose multiple binnings by concatenating the resulting embeddings before passing them to the model. This does not only allow us to use both local and global binnings, but also enables us to provide the model with multi-scale inputs by passing binnings with varying bin sizes and bin edges. 

\subsection{Output Representations}

In terms of modeling the output distribution $p(z_t|h_t)$, we compare three different approaches: a parametric distribution, in particular a Student-$t$ distribution ($\texttt{st}$) applied to the raw target values $z_{i, t}$, but where the mean and variance are scaled by the mean value computed on the conditioning range for each time series (an 
approach used for example in~\cite{flunkert2019deepar}); the piecewise-linear spline quantile function approach of \citet{gasthaus2019probabilistic} ($\texttt{plqs}$); and a categorical distribution applied to the binned values obtained through one of the binning strategies described above, where the final forecasts are obtained by applying the reconstruction function $s(\cdot)$ to samples from the predictive distribution.  

%
%
%
%
%

\subsection{Models}
\label{sec:models}

As part of this study we consider three different base models which we combine with the aforementioned input and output transformations: a variant of the WaveNet CNN architecture \citep{oord2016wavenet}, the DeepAR RNN architecture \citep{flunkert2019deepar}, and a basic two-layer feed-forward model.

\subsubsection{WaveNet (CNN)}
The WaveNet \citep{oord2016wavenet} architecture is an auto-regressive convolutional neural network which uses 1-dimensional dilated causal convolutions. While originally developed for speech synthesis, it has also been shown to be effective for time series forecasting \citep{borovykh2017}. More generally, CNN-based architectures using dilated causal convolutions have been demonstrated promising performance on a variety of sequence modeling tasks \citep{bai2018}. The specific model used in the experiments is the original WaveNet architecture as described in \citep{oord2016wavenet}, but using only a single stack of dilated convolutions with exponentially increasing dilation factor.

\subsubsection{DeepAR (RNN)}
DeepAR \citep{flunkert2019deepar} is an auto-regressive recurrent neural network architecture designed for time series forecasting. At its core, DeepAR is a RNN consisting of LSTM cells, which additionally receive auto-regressive inputs in the form of lagged target values. 

\subsubsection{Simple Deep Neural Network (Feed-Forward)}

The simple feed-forward model is a deep neural network which directly maps the past input sequence to the parameters of a multi-step output distribution without any feedback loops or memory. 
The model used in the experiments is a simple, plain two-layer model with 40 hidden units per layer with ReLU activation function and no additional regularization tricks (e.g.\ dropout, batch-norm, weight decay). No additional features or lags are used as inputs to this model.

%
%

\section{Experiments}
\label{sec:exp}
\begin{table*}[p]
\caption{Results with a fixed input global relative binning with 1024 quantile bins and varying output representations.}
\vspace{-12pt}
\label{tab:main_bin_output}
\begin{center}  
\small
\renewcommand{\arraystretch}{1.0}
\begin{tabular}{cccccccc}
\toprule     
& & \multicolumn{2}{c}{WaveNet} & \multicolumn{2}{c}{DeepAR} & \multicolumn{2}{c}{FeedForw}\\
\cmidrule(lr){3-4}
\cmidrule(lr){5-6}
\cmidrule(lr){7-8}
Dataset & Output & Mean wQL & ND & Mean wQL & ND & Mean wQL & ND\\

\midrule
\midrule

\texttt{m4\_h}       
& ms               & 0.0988 ($\pm$ 0.0871)  & 0.1135 ($\pm$ 0.0940) & 0.0566 ($\pm$ 0.0096)  & 0.0676 ($\pm$ 0.0102) & 0.0407 ($\pm$ 0.0028)  & 0.0519 ($\pm$ 0.0015) \\
& ms-plqs          & 0.0453 ($\pm$ 0.0110)  & 0.0557 ($\pm$ 0.0106) & 0.1462 ($\pm$ 0.0257)  & 0.1618 ($\pm$ 0.0289) & NaN ($\pm$ NaN)  & NaN ($\pm$ NaN)\\
& grb(bin1024)     & \textbf{0.0371} ($\pm$ 0.0092)  & 0.0487 ($\pm$ 0.0132) & 0.0953 ($\pm$ 0.0176)  & 0.1071 ($\pm$ 0.0152) & 0.0428 ($\pm$ 0.0006)  & 0.0539 ($\pm$ 0.0010)\\
& grb(bin1024,iqF) & 0.1292 ($\pm$ 0.0083)  & 0.1518 ($\pm$ 0.0170) & 0.0779 ($\pm$ 0.0155)  & 0.0890 ($\pm$ 0.0120) & 0.0468 ($\pm$ 0.0007)  & 0.0588 ($\pm$ 0.0009)\\
& lab(bin1024)     & 0.0372 ($\pm$ 0.0029)  & \textbf{0.0463} ($\pm$ 0.0028) & 0.0979 ($\pm$ 0.0134)  & 0.1123 ($\pm$ 0.0177) & 0.0419 ($\pm$ 0.0004)  & 0.0528 ($\pm$ 0.0004)\\

\midrule

\texttt{m4\_d}       
& ms               & 0.0260 ($\pm$ 0.0030)  & 0.0321 ($\pm$ 0.0040) & 0.0282 ($\pm$ 0.0009)  & 0.0338 ($\pm$ 0.0012) & 0.0298 ($\pm$ 0.0001)  & 0.0304 ($\pm$ 0.0001) \\
& ms-plqs          & 0.0237 ($\pm$ 0.0013)  & 0.0289 ($\pm$ 0.0019) & 0.0300 ($\pm$ 0.0021)  & 0.0363 ($\pm$ 0.0030) & NaN ($\pm$ NaN)  & NaN ($\pm$ NaN)\\
& grb(bin1024)     & \textbf{0.0228} ($\pm$ 0.0004)  & \textbf{0.0280} ($\pm$ 0.0005) & 0.2134 ($\pm$ 0.0181)  & 0.2235 ($\pm$ 0.0214) & 0.0307 ($\pm$ 0.0001)  & 0.0353 ($\pm$ 0.0000)\\
& grb(bin1024,iqF) & 0.0530 ($\pm$ 0.0009)  & 0.0629 ($\pm$ 0.0012) & 0.2103 ($\pm$ 0.0184)  & 0.2187 ($\pm$ 0.0195) & 0.0283 ($\pm$ 0.0000)  & 0.0330 ($\pm$ 0.0001)\\
& lab(bin1024)     & 0.0359 ($\pm$ 0.0003)  & 0.0412 ($\pm$ 0.0002) & 0.1675 ($\pm$ 0.0033)  & 0.2132 ($\pm$ 0.0017) & 0.0316 ($\pm$ 0.0000)  & 0.0367 ($\pm$ 0.0001)\\

\midrule

\texttt{m4\_w}       
& ms               & 0.0547 ($\pm$ 0.0039)  & 0.0686 ($\pm$ 0.0049) & 0.0455 ($\pm$ 0.0016)  & \textbf{0.0565} ($\pm$ 0.0026) & 0.0705 ($\pm$ 0.0002)  & 0.0811 ($\pm$ 0.0002) \\
& ms-plqs          & 0.0502 ($\pm$ 0.0038)  & 0.0626 ($\pm$ 0.0045) & 0.0477 ($\pm$ 0.0031)  & 0.0570 ($\pm$ 0.0056) & NaN ($\pm$ NaN)  & NaN ($\pm$ NaN)\\
& grb(bin1024)     & \textbf{0.0447} ($\pm$ 0.0016)  & 0.0569 ($\pm$ 0.0021) & 0.1746 ($\pm$ 0.0142)  & 0.1947 ($\pm$ 0.0133) & 0.0725 ($\pm$ 0.0002)  & 0.0851 ($\pm$ 0.0002)\\
& grb(bin1024,iqF) & 0.0641 ($\pm$ 0.0024)  & 0.0799 ($\pm$ 0.0029) & 0.1724 ($\pm$ 0.0150)  & 0.1899 ($\pm$ 0.0133) & 0.0707 ($\pm$ 0.0002)  & 0.0832 ($\pm$ 0.0001)\\
& lab(bin1024)     & 0.0623 ($\pm$ 0.0013)  & 0.0770 ($\pm$ 0.0018) & 0.2140 ($\pm$ 0.0036)  & 0.2446 ($\pm$ 0.0028) & 0.0764 ($\pm$ 0.0002)  & 0.0885 ($\pm$ 0.0002)\\

\midrule

\texttt{m4\_m}       
& ms               & 0.1313 ($\pm$ 0.0046)  & 0.1576 ($\pm$ 0.0049) & 0.1376 ($\pm$ 0.0123)  & 0.1639 ($\pm$ 0.028) & 0.1227 ($\pm$ 0.0009)  & 0.1589 ($\pm$ 0.0007) \\
& ms-plqs          & 0.1378 ($\pm$ 0.0013)  & 0.1595 ($\pm$ 0.0028) & 0.1471 ($\pm$ 0.0149)  & 0.1648 ($\pm$ 0.0062) & NaN ($\pm$ NaN)  & NaN ($\pm$ NaN)\\
& grb(bin1024)     & \textbf{0.1177} ($\pm$ 0.0031)  & \textbf{0.1447} ($\pm$ 0.0030) & 0.1755 ($\pm$ 0.0223)  & 0.2046 ($\pm$ 0.0048) & 0.1273 ($\pm$ 0.0004)  & 0.1466 ($\pm$ 0.0003)\\
& grb(bin1024,iqF) & 0.1429 ($\pm$ 0.0034)  & 0.1749 ($\pm$ 0.0054) & 0.1727 ($\pm$ 0.0220)  & 0.2049 ($\pm$ 0.0020) & 0.1268 ($\pm$ 0.009)  & 0.1454 ($\pm$ 0.0007)\\
& lab(bin1024)     & 0.1507 ($\pm$ 0.0003)  & 0.1819 ($\pm$ 0.0022) & 0.1931 ($\pm$ 0.0254)  & 0.2257 ($\pm$ 0.0089) & 0.1231 ($\pm$ 0.0006)  & 0.1470 ($\pm$ 0.0006)\\

\midrule

\texttt{m4\_q}       
& ms               & 0.0936 ($\pm$ 0.0032)  & 0.1148 ($\pm$ 0.0036) & 0.1067 ($\pm$ 0.0039)  & 0.1299 ($\pm$ 0.0047) & 0.1097 ($\pm$ 0.0007)  & 0.1299 ($\pm$ 0.0002) \\
& ms-plqs          & 0.0987 ($\pm$ 0.0028)  & 0.1188 ($\pm$ 0.0035) & 0.1267 ($\pm$ 0.0117)  & 0.1439 ($\pm$ 0.0108) & NaN ($\pm$ NaN)  & NaN ($\pm$ NaN)\\
& grb(bin1024)     & \textbf{0.0908} ($\pm$ 0.0015)  & \textbf{0.1126} ($\pm$ 0.0018) & 0.1673 ($\pm$ 0.0060)  & 0.1903 ($\pm$ 0.0044) & 0.1146 ($\pm$ 0.0011)  & 0.1318 ($\pm$ 0.0003)\\
& grb(bin1024,iqF) & 0.0998 ($\pm$ 0.0014)  & 0.1237 ($\pm$ 0.0017) & 0.1591 ($\pm$ 0.0092)  & 0.1819 ($\pm$ 0.0073) & 0.1131 ($\pm$ 0.0013)  & 0.1291 ($\pm$ 0.0003)\\
& lab(bin1024)     & 0.1195 ($\pm$ 0.0019)  & 0.1412 ($\pm$ 0.0022) & 0.1647 ($\pm$ 0.0103)  & 0.1980 ($\pm$ 0.0087) & 0.1197 ($\pm$ 0.0004)  & 0.1374 ($\pm$ 0.0002)\\

\midrule

\texttt{m4\_y}       
& ms               & \textbf{0.1235} ($\pm$ 0.0030)  & \textbf{0.1476} ($\pm$ 0.0032) & 0.1733 ($\pm$ 0.0073)  & 0.1940 ($\pm$ 0.0072) & 0.1262 ($\pm$ 0.0014)  & 0.1497 ($\pm$ 0.0014) \\
& ms-plqs          & 0.1271 ($\pm$ 0.0033)  & 0.1486 ($\pm$ 0.0039) & 0.1758 ($\pm$ 0.0200)  & 0.1973 ($\pm$ 0.0206) & NaN ($\pm$ NaN)  & NaN ($\pm$ NaN)\\
& grb(bin1024)     & 0.1538 ($\pm$ 0.0112)  & 0.1860 ($\pm$ 0.0140) & 0.2765 ($\pm$ 0.0076)  & 0.3073 ($\pm$ 0.0110) & 0.2131 ($\pm$ 0.0008)  & 0.2295 ($\pm$ 0.0002) \\
& grb(bin1024,iqF) & 0.1407 ($\pm$ 0.0116)  & 0.1712 ($\pm$ 0.0122) & 0.3001 ($\pm$ 0.0148)  & 0.3264 ($\pm$ 0.0124) & 0.2100 ($\pm$ 0.0009)  & 0.2254 ($\pm$ 0.0002)\\
& lab(bin1024)     & 0.2024 ($\pm$ 0.0110)  & 0.2237 ($\pm$ 0.0169) & 0.2596 ($\pm$ 0.0081)  & 0.3034 ($\pm$ 0.0135) & 0.2300 ($\pm$ 0.0003)  & 0.2399 ($\pm$ 0.0006)\\

\midrule

\texttt{elec}       
& ms               & 0.0610 ($\pm$ 0.0018)  & 0.0774 ($\pm$ 0.0028) & 0.0551 ($\pm$ 0.0011)  & 0.0678 ($\pm$ 0.0016) & 0.0668 ($\pm$ 0.0010)  & 0.0826 ($\pm$ 0.0017) \\
& ms-plqs          & 0.0540 ($\pm$ 0.0028)  & 0.0681 ($\pm$ 0.0036) & 0.0582 ($\pm$ 0.0029)  & 0.0707 ($\pm$ 0.0030) & NaN ($\pm$ NaN)  & NaN ($\pm$ NaN)\\
& grb(bin1024)     & \textbf{0.0475} ($\pm$ 0.0016)  & \textbf{0.0588} ($\pm$ 0.0024) & 0.0647 ($\pm$ 0.0018)  & 0.0821 ($\pm$ 0.0020) & 0.0677 ($\pm$ 0.0013)  & 0.0841 ($\pm$ 0.0020)\\
& grb(bin1024,iqF) & 0.0512 ($\pm$ 0.0015)  & 0.0641 ($\pm$ 0.0013) & 0.0712 ($\pm$ 0.0064)  & 0.0905 ($\pm$ 0.0088) & 0.0661 ($\pm$ 0.0007)  & 0.0817 ($\pm$ 0.0008)\\
& lab(bin1024)     & 0.0527 ($\pm$ 0.0009)  & 0.0647 ($\pm$ 0.0007) & 0.0791 ($\pm$ 0.0001)  & 0.1013 ($\pm$ 0.0002) & 0.0671 ($\pm$ 0.0010)  & 0.0793 ($\pm$ 0.0012)\\

\midrule

\texttt{traff}       
& ms               & 0.1437 ($\pm$ 0.0044)  & 0.1721 ($\pm$ 0.0050) & \textbf{0.1185} ($\pm$ 0.0089)  & \textbf{0.1401} ($\pm$ 0.0014) & 0.2111 ($\pm$ 0.0008)  & 0.2527 ($\pm$ 0.0008) \\
& ms-plqs          & 0.1237 ($\pm$ 0.0034)  & 0.1510 ($\pm$ 0.0040) & 0.1369 ($\pm$ 0.0045)  & 0.1680 ($\pm$ 0.0016) & 0.3185 ($\pm$ 0.1331)  & 0.3849 ($\pm$ 0.1646)\\
& grb(bin1024)     & 0.1209 ($\pm$ 0.0016)  & 0.1455 ($\pm$ 0.0019) & 0.1883 ($\pm$ 0.0012)  & 0.2323 ($\pm$ 0.0005) & 0.2218 ($\pm$ 0.0010)  & 0.0252 ($\pm$ 0.0015)\\
& grb(bin1024,iqF) & 0.1229 ($\pm$ 0.0017)  & 0.1495 ($\pm$ 0.0022) & 0.1835 ($\pm$ 0.0034)  & 0.2245 ($\pm$ 0.0012) & 0.2210 ($\pm$ 0.0017)  & 0.0341 ($\pm$ 0.0050)\\
& lab(bin1024)     & 0.1261 ($\pm$ 0.0006)  & 0.1536 ($\pm$ 0.0007) & 0.1632 ($\pm$ 0.0051)  & 0.2005 ($\pm$ 0.0028) & 0.3820 ($\pm$ 0.0832)  & 0.0206 ($\pm$ 0.0058)\\

\midrule

\texttt{wiki}       
& ms               & 0.2204 ($\pm$ 0.0021)  & 0.2480 ($\pm$ 0.0027) & 0.2284 ($\pm$ 0.0012)  & 0.2577 ($\pm$ 0.0012) & 0.2721 ($\pm$ 0.0015)  & 0.3349 ($\pm$ 0.0014) \\
& ms-plqs          & 0.2336 ($\pm$ 0.0097)  & 0.2654 ($\pm$ 0.0118) & 0.2305 ($\pm$ 0.0040)  & 0.2561 ($\pm$ 0.0033) & NaN ($\pm$ NaN)  & NaN ($\pm$ NaN)\\
& grb(bin1024)     & \textbf{0.2156} ($\pm$ 0.0020)  & \textbf{0.2439} ($\pm$ 0.0021) & 0.8465 ($\pm$ 0.0199)  & 0.9459 ($\pm$ 0.0270) & 0.2930 ($\pm$ 0.0013)  & 0.3316 ($\pm$ 0.0010)\\
& grb(bin1024,iqF) & 0.2224 ($\pm$ 0.0037)  & 0.2524 ($\pm$ 0.0044) & 0.7919 ($\pm$ 0.0019)  & 0.9086 ($\pm$ 0.0076) & 0.2961 ($\pm$ 0.0015)  & 0.3346 ($\pm$ 0.0009)\\
& lab(bin1024)     & 0.2564 ($\pm$ 0.0016)  & 0.2901 ($\pm$ 0.0021) & 0.6996 ($\pm$ 0.0009)  & 0.8200 ($\pm$ 0.0007) & 0.2540 ($\pm$ 0.0004)  & 0.2858 ($\pm$ 0.0005)\\

\bottomrule
\end{tabular}
\end{center}
\end{table*}

\begin{table*}[p]
\caption{Results with a fixed output global relative binning with 1024 quantile bins and varying input representations.}
\vspace{-12pt}
\label{tab:main_bin_input}
\begin{center}  
\small
\renewcommand{\arraystretch}{0.85}
\begin{tabular}{cccccccc}
\toprule     
& & \multicolumn{2}{c}{WaveNet} & \multicolumn{2}{c}{DeepAR} & \multicolumn{2}{c}{FeedForw}\\
\cmidrule(lr){3-4}
\cmidrule(lr){5-6}
\cmidrule(lr){7-8}
Dataset & Input & Mean wQL & ND & Mean wQL & ND & Mean wQL & ND\\

\midrule
\midrule

\texttt{m4\_h}       
& ms               & 0.0391 ($\pm$ 0.0057)  & 0.0506 ($\pm$ 0.0083) & 0.0931 ($\pm$ 0.0093)  & 0.1066 ($\pm$ 0.0090) & 0.0463 ($\pm$ 0.0005)  & 0.0588 ($\pm$ 0.0011) \\
& lab(bin1024)     & 0.0577 ($\pm$ 0.0075)  & 0.0736 ($\pm$ 0.0099) & 0.1114 ($\pm$ 0.0078)  & 0.1255 ($\pm$ 0.0101) & 0.0517 ($\pm$ 0.0035)  & 0.0643 ($\pm$ 0.0034)\\
& pit(bin1024)     & \textbf{0.0296} ($\pm$ 0.0001)  & \textbf{0.0370} ($\pm$ 0.0002) & 0.0902 ($\pm$ 0.0089)  & 0.1120 ($\pm$ 0.0095) & 0.0721 ($\pm$ 0.0392)  & 0.0912 ($\pm$ 0.0491)\\
& hyb(16,128,1024) & 0.0375 ($\pm$ 0.0009)  & 0.0504 ($\pm$ 0.0003) & 0.1020 ($\pm$ 0.0057)  & 0.1189 ($\pm$ 0.0109) & 0.0435 ($\pm$ 0.0004)  & 0.0549 ($\pm$ 0.0006)\\
& hyb(grb,lab)     & 0.0369 ($\pm$ 0.0061)  & 0.0475 ($\pm$ 0.0089) & 0.1057 ($\pm$ 0.0088)  & 0.1201 ($\pm$ 0.0110) & 0.0421 ($\pm$ 0.0017)  & 0.0537 ($\pm$ 0.0021)\\

\midrule

\texttt{m4\_d}       
& ms               & 0.0315 ($\pm$ 0.0057)  & 0.0378 ($\pm$ 0.0065) & 0.2128 ($\pm$ 0.0182)  & 0.2216 ($\pm$ 0.0188) & 0.0305 ($\pm$ 0.0000)  & 0.0352 ($\pm$ 0.0000) \\
& lab(bin1024)     & 0.0317 ($\pm$ 0.0007)  & 0.0369 ($\pm$ 0.0008) & 0.2189 ($\pm$ 0.0124)  & 0.2244 ($\pm$ 0.0130) & 0.0305 ($\pm$ 0.0000)  & 0.0352 ($\pm$ 0.0001)\\
& pit(bin1024)     & 0.0286 ($\pm$ 0.0053)  & 0.0345 ($\pm$ 0.0061) & 0.2204 ($\pm$ 0.0144)  & 0.2283 ($\pm$ 0.0141) & 0.0305 ($\pm$ 0.0002)  & 0.0352 ($\pm$ 0.0000)\\
& hyb(16,128,1024) & \textbf{0.0227} ($\pm$ 0.0003)  & \textbf{0.0278} ($\pm$ 0.0004) & 0.2196 ($\pm$ 0.0137)  & 0.2267 ($\pm$ 0.0135) & 0.0306 ($\pm$ 0.0001)  & 0.0353 ($\pm$ 0.0001)\\
& hyb(grb,lab)     & 0.0272 ($\pm$ 0.0004)  & 0.0318 ($\pm$ 0.0003) & 0.2222 ($\pm$ 0.0156)  & 0.2301 ($\pm$ 0.0157) & 0.0307 ($\pm$ 0.0001)  & 0.0353 ($\pm$ 0.0000)\\

\midrule

\texttt{m4\_w}       
& ms               & 0.0848 ($\pm$ 0.0327)  & 0.1026 ($\pm$ 0.0371) & 0.1651 ($\pm$ 0.0113)  & 0.1830 ($\pm$ 0.0111) & 0.0750 ($\pm$ 0.0005)  & 0.0839 ($\pm$ 0.0001) \\
& lab(bin1024)     & 0.1061 ($\pm$ 0.0023)  & 0.1244 ($\pm$ 0.0033) & 0.1838 ($\pm$ 0.0070)  & 0.1995 ($\pm$ 0.0083) & 0.0760 ($\pm$ 0.0002)  & 0.0834 ($\pm$ 0.0001)\\
& pit(bin1024)     & 0.0467 ($\pm$ 0.0022)  & 0.0585 ($\pm$ 0.0028) & 0.1884 ($\pm$ 0.0099)  & 0.2082 ($\pm$ 0.0103) & 0.0724 ($\pm$ 0.0005)  & 0.0848 ($\pm$ 0.0004)\\
& hyb(16,128,1024) & \textbf{0.0443} ($\pm$ 0.0010)  & \textbf{0.0561} ($\pm$ 0.0014) & 0.1792 ($\pm$ 0.0047)  & 0.1980 ($\pm$ 0.0042) & 0.0723 ($\pm$ 0.0002)  & 0.0854 ($\pm$ 0.0002)\\
& hyb(grb,lab)     & 0.0500 ($\pm$ 0.0012)  & 0.0627 ($\pm$ 0.0015) & 0.1815 ($\pm$ 0.0072)  & 0.1975 ($\pm$ 0.0074) & 0.0719 ($\pm$ 0.0003)  & 0.0849 ($\pm$ 0.0002)\\

\midrule

\texttt{m4\_m}       
& ms               & 0.1373 ($\pm$ 0.0143)  & 0.1655 ($\pm$ 0.0137) & 0.2080 ($\pm$ 0.0102)  & 0.2412 ($\pm$ 0.0098) & 0.1392 ($\pm$ 0.0009)  & 0.1470 ($\pm$ 0.0000) \\
& lab(bin1024)     & 0.2055 ($\pm$ 0.0021)  & 0.2136 ($\pm$ 0.0012) & 0.2395 ($\pm$ 0.0154)  & 0.2891 ($\pm$ 0.0101) & 0.1396 ($\pm$ 0.0005)  & 0.1463 ($\pm$ 0.0001)\\
& pit(bin1024)     & 0.1213 ($\pm$ 0.0024)  & 0.1481 ($\pm$ 0.0029) & 0.1921 ($\pm$ 0.0097)  & 0.2287 ($\pm$ 0.0084) & 0.1332 ($\pm$ 0.0049)  & 0.1462 ($\pm$ 0.0009)\\
& hyb(16,128,1024) & \textbf{0.1187} ($\pm$ 0.0037)  & 0.1463 ($\pm$ 0.0046) & 0.1944 ($\pm$ 0.0098)  & 0.2294 ($\pm$ 0.0057) & 0.1267 ($\pm$ 0.0023)  & 0.1459 ($\pm$ 0.0001)\\
& hyb(grb,lab)     & 0.1206 ($\pm$ 0.0010)  & 0.1468 ($\pm$ 0.0008) & 0.2018 ($\pm$ 0.0105)  & 0.2388 ($\pm$ 0.0083) & 0.1264 ($\pm$ 0.0014)  & \textbf{0.1454} ($\pm$ 0.0002)\\

\midrule

\texttt{m4\_q}       
& ms               & 0.1272 ($\pm$ 0.0006)  & 0.1488 ($\pm$ 0.0003) & 0.1507 ($\pm$ 0.0037)  & 0.1698 ($\pm$ 0.0021) & 0.1256 ($\pm$ 0.0009)  & 0.1501 ($\pm$ 0.0008) \\
& lab(bin1024)     & 0.1299 ($\pm$ 0.0017)  & 0.1486 ($\pm$ 0.0013) & 0.1689 ($\pm$ 0.0025)  & 0.1861 ($\pm$ 0.0016) & 0.1174 ($\pm$ 0.0011)  & 0.1320 ($\pm$ 0.0004)\\
& pit(bin1024)     & 0.1278 ($\pm$ 0.0014)  & 0.1488 ($\pm$ 0.0002) & 0.1748 ($\pm$ 0.0028)  & 0.1958 ($\pm$ 0.0021) & 0.1180 ($\pm$ 0.0011)  & 0.1324 ($\pm$ 0.0002)\\
& hyb(16,128,1024)& \textbf{0.0893} ($\pm$ 0.0011)  & \textbf{0.1108} ($\pm$ 0.0012) & 0.1743 ($\pm$ 0.0052)  & 0.1972 ($\pm$ 0.0035) & 0.1152 ($\pm$ 0.0019)  & 0.1314 ($\pm$ 0.0007)\\
& hyb(grb,lab)     & 0.1137 ($\pm$ 0.0032)  & 0.1372 ($\pm$ 0.0034) & 0.1722 ($\pm$ 0.0029)  & 0.1974 ($\pm$ 0.0008) & 0.1152 ($\pm$ 0.0021)  & 0.1308 ($\pm$ 0.0007)\\

\midrule

\texttt{m4\_y}       
& ms               & \textbf{0.1308} ($\pm$ 0.0039)  & \textbf{0.1562} ($\pm$ 0.0034) & 0.2663 ($\pm$ 0.0177)  & 0.2907 ($\pm$ 0.0123) & 0.2162 ($\pm$ 0.0016)  & 0.2326 ($\pm$ 0.0008) \\
& lab(bin1024)     & 0.2812 ($\pm$ 0.0144)  & 0.3171 ($\pm$ 0.0094) & 0.3062 ($\pm$ 0.0140)  & 0.3248 ($\pm$ 0.0085) & 0.2143 ($\pm$ 0.0004)  & 0.2309 ($\pm$ 0.0002)\\
& pit(bin1024)     & 0.1844 ($\pm$ 0.0523)  & 0.2202 ($\pm$ 0.0621) & 0.3058 ($\pm$ 0.0077)  & 0.3280 ($\pm$ 0.0086) & 0.2151 ($\pm$ 0.0019)  & 0.2324 ($\pm$ 0.0022)\\
& hyb(16,128,1024) & 0.1337 ($\pm$ 0.0033)  & 0.1618 ($\pm$ 0.0045) & 0.2925 ($\pm$ 0.0028)  & 0.3219 ($\pm$ 0.0024) & 0.2129 ($\pm$ 0.0006)  & 0.2295 ($\pm$ 0.0001)\\
& hyb(grb,lab)     & 0.2065 ($\pm$ 0.0149)  & 0.2505 ($\pm$ 0.0195) & 0.3184 ($\pm$ 0.0050)  & 0.3576 ($\pm$ 0.0052) & 0.2235 ($\pm$ 0.0068)  & 0.2388 ($\pm$ 0.0020)\\

\midrule

\texttt{elec}       
& ms               & 0.0501 ($\pm$ 0.0010)  & 0.0607 ($\pm$ 0.0017) & 0.0732 ($\pm$ 0.0007)  & 0.0923 ($\pm$ 0.0004) & 0.0800 ($\pm$ 0.0038)  & 0.1004 ($\pm$ 0.0056) \\
& lab(bin1024)     & 0.1389 ($\pm$ 0.0070)  & 0.1677 ($\pm$ 0.0096) & 0.0986 ($\pm$ 0.0023)  & 0.1107 ($\pm$ 0.0067) & 0.1269 ($\pm$ 0.0033)  & 0.1632 ($\pm$ 0.0034)\\
& pit(bin1024)     & 0.0484 ($\pm$ 0.0010)  & 0.0598 ($\pm$ 0.0015) & 0.4210 ($\pm$ 0.1192)  & 0.4924 ($\pm$ 0.1078) & 0.0705 ($\pm$ 0.0026)  & 0.0875 ($\pm$ 0.0041)\\
& hyb(16,128,1024) & 0.0495 ($\pm$ 0.0004)  & 0.0612 ($\pm$ 0.0007) & 0.1143 ($\pm$ 0.0028)  & 0.1339 ($\pm$ 0.0055) & 0.0678 ($\pm$ 0.0018)  & 0.0843 ($\pm$ 0.0026)\\
& hyb(grb,lab)     & \textbf{0.0472} ($\pm$ 0.0005)  & \textbf{0.0585} ($\pm$ 0.0005) & 0.1528 ($\pm$ 0.0072)  & 0.1801 ($\pm$ 0.0089) & 0.0687 ($\pm$ 0.0009)  & 0.0856 ($\pm$ 0.0011)\\

\midrule

\texttt{traff}       
& ms               & 0.1251 ($\pm$ 0.0013)  & 0.1507 ($\pm$ 0.0013) & 0.1974 ($\pm$ 0.0088)  & 0.2423 ($\pm$ 0.0339) & 0.2280 ($\pm$ 0.0005)  & 0.0287 ($\pm$ 0.0021) \\
& lab(bin1024)     & 0.2571 ($\pm$ 0.0174)  & 0.3200 ($\pm$ 0.0246) & 0.2535 ($\pm$ 0.0246)  & 0.3131 ($\pm$ 0.0632) & 0.2456 ($\pm$ 0.0010)  & 0.0070 ($\pm$ 0.0021)\\
& pit(bin1024)     & 0.1275 ($\pm$ 0.0008)  & 0.1539 ($\pm$ 0.0010) & 0.5953 ($\pm$ 0.1299)  & 0.7266 ($\pm$ 0.2040) & 0.2258 ($\pm$ 0.0017)  & 0.0254 ($\pm$ 0.0037)\\
& hyb(16,128,1024) & \textbf{0.1242} ($\pm$ 0.0008)  & \textbf{0.1498} ($\pm$ 0.0009) & 0.1886 ($\pm$ 0.0072)  & 0.2316 ($\pm$ 0.0290) & 0.2184 ($\pm$ 0.0011)  & 0.0249 ($\pm$ 0.0016)\\
& hyb(grb,lab)     & 0.1245 ($\pm$ 0.0011)  & 0.1505 ($\pm$ 0.0018) & 0.1885 ($\pm$ 0.0091)  & 0.2315 ($\pm$ 0.0387) & 0.2182 ($\pm$ 0.0007)  & 0.0217 ($\pm$ 0.0028)\\

\midrule

\texttt{wiki}       
& ms               & 0.2183 ($\pm$ 0.0028)  & 0.2472 ($\pm$ 0.0033) & 0.8156 ($\pm$ 0.0176)  & 0.9170 ($\pm$ 0.0234) & 0.3027 ($\pm$ 0.0007)  & 0.3381 ($\pm$ 0.0008) \\
& lab(bin1024)     & 0.3071 ($\pm$ 0.0030)  & 0.3478 ($\pm$ 0.0026) & 0.8143 ($\pm$ 0.0112)  & 0.9115 ($\pm$ 0.0130) & 0.3066 ($\pm$ 0.0005)  & 0.3408 ($\pm$ 0.0004)\\
& pit(bin1024)     & 0.2177 ($\pm$ 0.0043)  & 0.2465 ($\pm$ 0.0047) & 0.9238 ($\pm$ 0.2095)  & 0.9981 ($\pm$ 0.3049) & 0.2935 ($\pm$ 0.0014)  & 0.3304 ($\pm$ 0.0014)\\
& hyb(16,128,1024) & \textbf{0.2163} ($\pm$ 0.0017)  & \textbf{0.2447} ($\pm$ 0.0019) & 0.8140 ($\pm$ 0.0129)  & 0.9075 ($\pm$ 0.0090) & 0.2927 ($\pm$ 0.0012)  & 0.3311 ($\pm$ 0.0008)\\
& hyb(grb,lab)     & 0.2342 ($\pm$ 0.0035)  & 0.2631 ($\pm$ 0.0037) & 0.8191 ($\pm$ 0.0150)  & 0.9238 ($\pm$ 0.0198) & 0.2931 ($\pm$ 0.0012)  & 0.3316 ($\pm$ 0.0011)\\

\bottomrule
\end{tabular}
\end{center}
\end{table*}

\begin{figure*}[t]
	\vspace{20pt}
    \centering
    \begin{subfigure}[t]{0.30\textwidth}
        \centering
        \includegraphics[width=.98\linewidth]{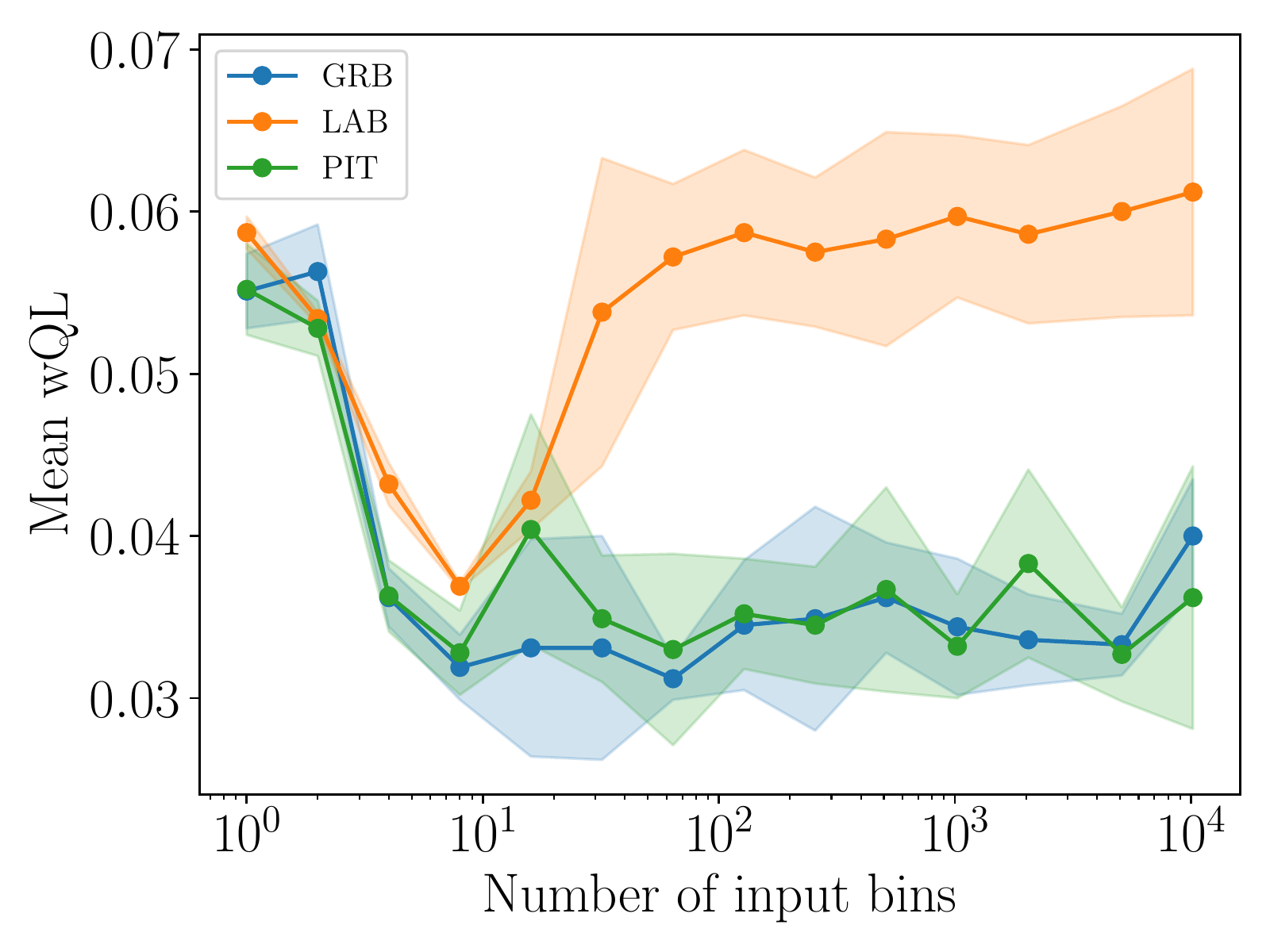}
        \caption{Performance effects of varying input resolutions with respect to a fixed global relative output with 1024 quantile bins. Although LAB does first improve and then deteriorate in performance, we see that the number of input bins does play a lesser role than the specified output distribution.}
    \end{subfigure}%
    \hspace{10pt}
    \begin{subfigure}[t]{0.30\textwidth}
        \centering
        \includegraphics[width=.98\linewidth]{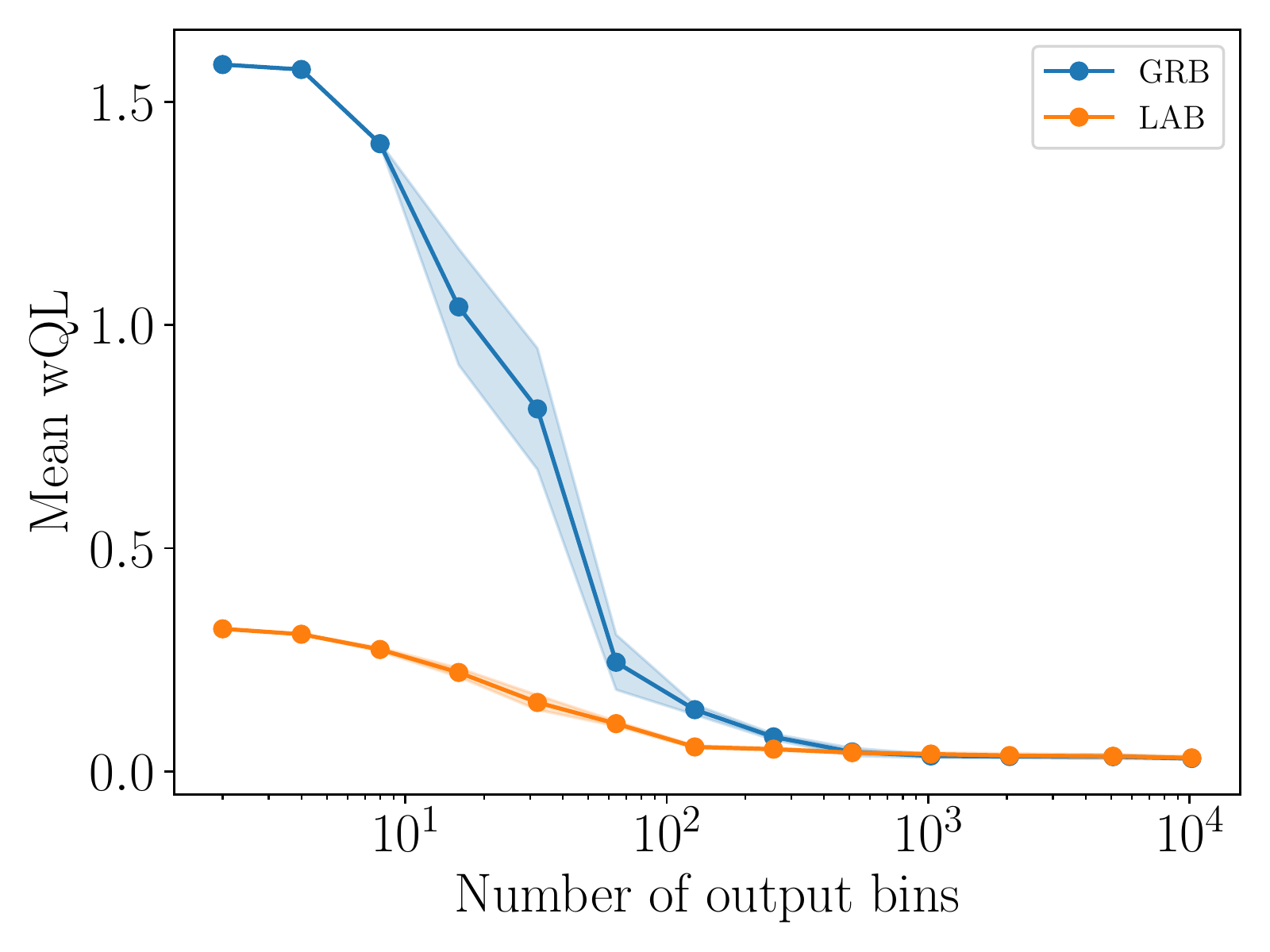}
        \caption{Performance effects of varying output resolutions with respect to a fixed global relative input with 1024 quantile bins. It is clearly visible that the chosen output representation plays a key role and that increasing the number of output bins improves performance.}
    \end{subfigure}
    \hspace{10pt}
    \begin{subfigure}[t]{0.30\textwidth}
        \centering
        \includegraphics[width=.98\linewidth]{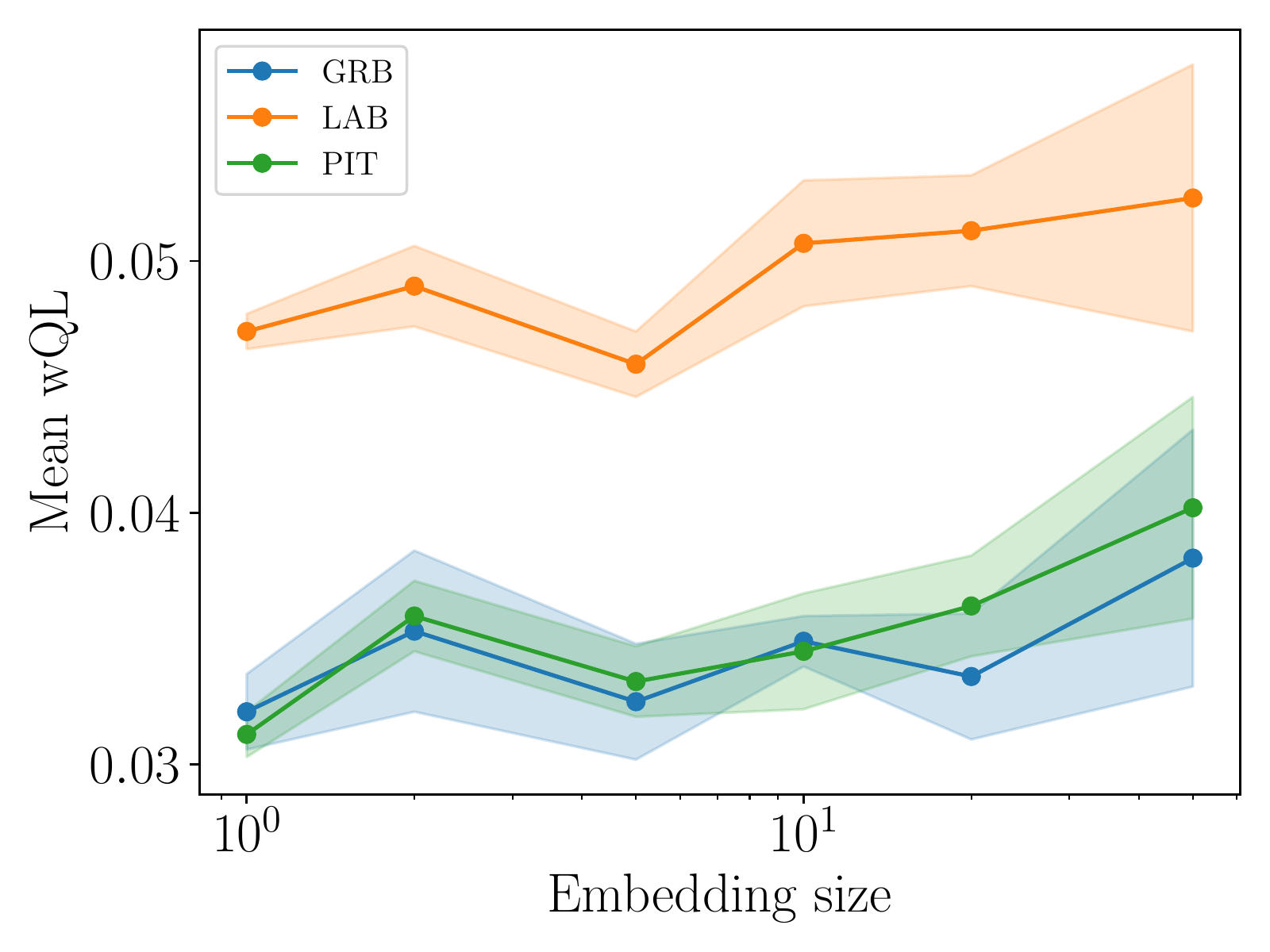}
        \caption{Performance effects of varying embedding sizes given a fixed global relative output with 1024 quantile bins and fixed 1024 input bins across different input representations. Similar to the number of input bins, the embedding size does not play a major role w.r.t model performance.}
    \end{subfigure}
 
    \caption{Insights into the performance effects incurred by altering the input/output bins, as well as the embedding size.}
    \label{fig:binning_insights}
\end{figure*}

Our experiments were conducted using GluonTS \citep{alexandrov2019gluonts}, a Python toolkit for probabilistic time series modeling using deep learning-based models
based on~\cite{chen2015mxnet}. GluonTS 
provides implementations of the models introduced in Section \ref{sec:models}, and we extended it with the ability to flexibly specify the input representations and output distributions. We trained all models on the data sets from the \texttt{m4} forecasting competition (\texttt{m4\_hourly}, \texttt{m4\_daily},  \texttt{m4\_weekly}, \texttt{m4\_monthly}, \texttt{m4\_quarterly}, \texttt{m4\_yearly}) \citep{makridakis2018m4}, on the \texttt{electricity} and \texttt{traffic} datasets \citep{Dua:2019}, and on a sample of daily page hits of Wikipedia subpages, \texttt{wiki10k}, using various combinations of representations and report mean error metrics and standard deviation over 10 random runs per configuration. Specifically, we investigated the performance effect of fixing the input representation while varying the output representation and vice versa and also examined the effects of the binning and embedding resolutions in detail. Although we do report results for \texttt{m4\_hourly}, we note that we specifically used this dataset for tuning hyper-parameters and for generating deeper insights on representation performance.

Both the \texttt{FeedForw} and \texttt{WaveNet} model were trained using the Adam optimizer with a decaying initial learning rate of $10^{-2}$ (decay rate \nicefrac{1}{2}) in batches of $32$ samples over $150$ epochs (where one epoch consists of $50$ batches) on a \texttt{p2.xlarge} instance (NVIDIA K80 GPU) on Amazon Web Services. \texttt{DeepAR} follows the same setting but starts with an initial learning rate of $10^{-3}$ and is trained over $100$ epochs. All models with the exception of \texttt{FeedForw} make use of supplementary covariates $\bm{x}$ encoding date-dependent features in the form of dummy variables in addition to time series target values $z$. Moreover, \texttt{DeepAR} further utilizes lagged values at varying frequencies (hourly, daily, weekly, etc.) for quicker convergence as it allows the model to pick up highly periodic patterns more easily.

By default, binnings are assumed to be quantile-based, utilize $B=1024$ bins, and are embedded in a $E=\sqrt[4]{B}$ dimensional space \cite{featurecolumns} before being fed into the model. Initially, we also experimented with linear binning, but found the quantile binnings to be generally more reliable. %
When using quantile splines on the output, we default to a resolution of $20$ knots.

We report predictive performance in the form of two commonly-used accuracy metrics: To evaluate the quality of the predictive distributions we measure mean weighted quantile loss, which is an approximation to the continuous ranked probability score \citep{Matheson1976scoring, Gneiting2007strictly}. In particular, we compute,
\begin{equation*}
   \text{mean wQL} = \frac{
    \sum_{i, t}\frac{2}{|A|}\sum_{\alpha \in A}(\alpha - I[z_{i,t} < q_{i,t}(\alpha)])(z_{i,t} - q_{i,t}(\alpha))
    }
    {\sum_{i, t} |z_{i,t}|},
\end{equation*}
where $q_{i,t}(\alpha)$ is the $\alpha$-quantile of the predictive distribution for $z_{i,t}$,
and $A = \{0.1, 0.2, \ldots, 0.9\}$ is the set of quantile levels we evaluate.
To evaluate the point forecasting performance, we evaluate the normalized deviation (ND), which is equivalent to wQL evaluated only at the median, i.e. with $A = \{0.5\}$.

The main results are shown in Table \ref{tab:main_bin_output} (which shows the effect of varying the output representation but keeping the input fixed), Table \ref{tab:main_bin_input} (different discretizing input transformations while keeping the output fixed), and Table \ref{tab:main_scaling} (performance with scaled input/output but no binning).

We performed additional experiments using the WaveNet model on the \texttt{m4\_hourly} data set to better understand the effect of the various transformation hyperparameters (\texttt{grb} vs. \texttt{lab} vs. \texttt{pit}; number of bins used; embedding size). The results of these experiments are shown in Figure \ref{fig:binning_insights}.

\section{Discussion}
\label{sec:diss}

In the following we summarize the observations and conclusions from our experiments. 

\paragraph{Output Scaling versus Binning} Our main results show (cf. first column in Table \ref{tab:main_bin_output}) that in particular the \texttt{WaveNet} model substantially benefits from binned output representations when compared to real-valued, scaled outputs modeled through a parametric Student-$t$ distribution (\texttt{ms}) or using the quantile spline output (\texttt{ms-plqs}). In fact, WaveNet combined with global relative (quantile) binning for the input and output transformation (\texttt{grb(bin1024)}) almost always (\texttt{m4\_y} being the exception) outperforms all other combinations in our comparison across datasets. Interestingly, for \texttt{DeepAR} this effect is reversed (cf. Table \ref{tab:main_bin_output}, col. 2) and mean-scaled, non-binned output representations substantially outperform the discretized ones. \texttt{FeedForw} shows no clear advantage for either of the representations, but generally performs worse than either of the other models in their best configuration. These results underline our claim that input/output representations in general and output representations in particular can be equally (or even more) important for obtaining good predictive performance than choosing a particular model class, and more powerful models like WaveNet can be outperformed by simpler models like the feed forward model if the representations are not carefully chosen (e.g.\ on \texttt{m4\_h}, \texttt{FeedForw} with mean-scaled Student-$t$ output (\texttt{ms}) outperforms WaveNet with the same output, but is in turn outperformed by WaveNet with global-relative binning (\texttt{grb})).

\paragraph{Input Scaling vs. Binning} Table \ref{tab:main_bin_input} shows the performance of the models when the input representation is varied while the output representation is fixed (\texttt{grb}). Interestingly, while the hybrid binning (\texttt{hyb(16,128,1024)}) often performs well, there is no clear dominant strategy here that outperforms the others across datasets and/or models. However, the impact of the input transformation on the  performance is also less pronounced than for the output. One notable exception is local-absolute binning (\texttt{lab}) which often performs significantly worse than the other strategies in this setting. This, in combination with the insensitivity to the number of input bins shown in Figure \ref{fig:binning_insights} hints at models' ability to extract sufficient information from either of the binnings, even when they have low resolution. Further, as expected, the (\texttt{pit}) strategy, being the continuous analogue of (\texttt{grb}), performs on par with it, though (\texttt{grb}) appears to have a slight edge.

\paragraph{Binning resolution effects} Interestingly, we found that (cf. Figure \ref{fig:binning_insights}), given a fixed global relative binning on the \emph{output} with 1024 quantile bins, a surprisingly small number of input bins already suffices to achieve good predictive accuracy and, more so, that increasing the number of input bins does not significantly improve performance. In contrast, given a fixed global relative binning on the \emph{input} with 1024 quantile bins, increasing the number of bins on the output leads to steady improvements in performance. While the latter effect mostly expected due to the reconstruction loss incurred with a discretized output with less bins (cf.\ Figure \ref{fig:reconstr}), the former effect is more surprising and hints at the fact the the models learn to focus on coarse-grained effects in the input, rather than focussing on fine details (that would be lost with a smaller number of bins).

\paragraph{Embedding size effects} Since the embedding size, which is governed by a heuristic described in Section~\ref{sec:exp}, is dependent on the number of bins, we also explicitly assess the performance impact of varying the embedding size in isolation, keeping the other parameters fixed (Figure \ref{fig:binning_insights} c)). Similar to the results reported in Figure \ref{fig:binning_insights} a), we found that altering the embedding size while keeping the number of bins fixed does not significantly impact performance, and that a relatively small embedding size is sufficient.

\paragraph{Global versus Local Binning} We observed that the global relative binning strategy tends to work better than local absolute binning for the output. While the effect is small on some datasets, it is more pronounced on others (e.g. WaveNet on \texttt{m4\_m}, \texttt{m4\_q}, and \texttt{m4\_y} in Table \ref{tab:main_bin_output}). Note that (\texttt{grb}) is used for the input transformation here, so that there is a ``mismatch'' between the input and the output binning, which seems to be responsible for part of this effect. However, we performed additional experiment with (\texttt{lab}) input transformation (not shown) where this effect is somewhat alleviated, but does not vanish.

\paragraph{Hybrid versus Single Binning} We also analyzed whether hybrid binning strategies used as an input transformation can improve performance over a single binning. Specifically, we considered two different kinds of hybrid binnings: \texttt{hyb(16,128,1024)} which includes multiple global relative binnings at different resolutions and \texttt{hyb(grb,lab)} which combines a global relative and a local absolute binning. Our results show that the multi-scale hybrid binning does indeed improve performance in many instances and is in fact the best-performing method reported for many datasets if used in conjunction with the WaveNet. However, combining both local and global information does not consistently lead to improvements over the best performing method, but rather averages results reported for global relative inputs and local absolute inputs. 

\paragraph{Models} Overall, WaveNet does profit the most from the proposed binning strategies, while the FeedForw model does not show any meaningful gains from using binning. As already hinted at, while DeepAR can make effective use of input binnings, it demonstrates significantly worse performance when combined with a binned output representation. The reason for this is not yet clear and would benefit from further investigation.

\section{Related Work}
\label{sec:related}
\begin{table*}
\caption{Results with mean scaling on both inputs and outputs. This is the standard scaling setting in GluonTS \citep{alexandrov2019gluonts}.}
\vspace{-10pt}
\label{tab:main_scaling}
\begin{center}  
\small
\renewcommand{\arraystretch}{0.9}
\begin{tabular}{ccccccc}
\toprule     
& \multicolumn{2}{c}{WaveNet} & \multicolumn{2}{c}{DeepAR} & \multicolumn{2}{c}{FeedForw}\\
\cmidrule(lr){2-3}
\cmidrule(lr){4-5}
\cmidrule(lr){6-7}
Dataset & Mean wQL & ND & Mean wQL & ND & Mean wQL & ND\\

\midrule
\midrule

\texttt{m4\_h}       
& 0.1517 ($\pm$ 0.0904)  & 0.2008 ($\pm$ 0.1334) & 0.0533 ($\pm$ 0.0012)  & 0.0645 ($\pm$ 0.0009) & \textbf{0.0463} ($\pm$ 0.0010)  & \textbf{0.0580} ($\pm$ 0.0012) \\
\texttt{m4\_d}       
& 0.0334 ($\pm$ 0.0088)  & 0.0401 ($\pm$ 0.0102) & 0.0318 ($\pm$ 0.0029)  & 0.0384 ($\pm$ 0.0036) & \textbf{0.0247} ($\pm$ 0.0005)  & \textbf{0.0296} ($\pm$ 0.0008) \\
\texttt{m4\_w}       
& 0.0574 ($\pm$ 0.0036)  & 0.0716 ($\pm$ 0.0042) & \textbf{0.0460} ($\pm$ 0.0011)  & \textbf{0.0565} ($\pm$ 0.0012) & 0.0521 ($\pm$ 0.0006)  & 0.0614 ($\pm$ 0.0006) \\
\texttt{m4\_m}       
& 0.1481 ($\pm$ 0.0170)  & 0.1674 ($\pm$ 0.0152) & 0.1362 ($\pm$ 0.0089)  & 0.1480 ($\pm$ 0.0083) & \textbf{0.1159} ($\pm$ 0.0011)  & \textbf{0.1260} ($\pm$ 0.0023) \\
\texttt{m4\_q}       
& 0.0983 ($\pm$ 0.0019)  & 0.1196 ($\pm$ 0.0017) & 0.1030 ($\pm$ 0.0031)  & 0.1176 ($\pm$ 0.0027) & \textbf{0.0869} ($\pm$ 0.0010)  & \textbf{0.1030} ($\pm$ 0.0010) \\
\texttt{m4\_y}       
& \textbf{0.1236} ($\pm$ 0.0055)  & \textbf{0.1458} ($\pm$ 0.0057) & 0.1570 ($\pm$ 0.0088)  & 0.1757 ($\pm$ 0.0085) & 0.1262 ($\pm$ 0.0014)  & 0.1497 ($\pm$ 0.0014) \\
\texttt{elec}       
& 0.0724 ($\pm$ 0.0151)  & 0.0923 ($\pm$ 0.0194) & \textbf{0.0571} ($\pm$ 0.0012)  & \textbf{0.0695} ($\pm$ 0.0018) & 0.0649 ($\pm$ 0.0011)  & 0.0793 ($\pm$ 0.0015) \\
\texttt{traff}       
& 0.1450 ($\pm$ 0.0065)  & 0.1720 ($\pm$ 0.0073) & \textbf{0.1222} ($\pm$ 0.0077)  & \textbf{0.1456} ($\pm$ 0.0082) & 0.2144 ($\pm$ 0.0008)  & 0.2558 ($\pm$ 0.0009) \\
\texttt{wiki}       
& \textbf{0.2295} ($\pm$ 0.0063)  & \textbf{0.2601} ($\pm$ 0.0072) & 0.2378 ($\pm$ 0.0070)  & 0.2694 ($\pm$ 0.0091) & 0.2594 ($\pm$ 0.0026)  & 0.3030 ($\pm$ 0.0036) \\
\bottomrule
\end{tabular}
\end{center}
\end{table*}

The empirical study presented here is part of a growing amount of literature on neural forecasting approaches~\cite{smyl2018m4,wang2019deepfactors,laptev2017,fan2019multi,li2019,KDD19Extreme,KDD19Streaming}. While most prior art considers the probabilistic forecasting setting, some recent work has resorted to only providing point forecasts~\cite{LSTNet,nbeats}. For forecasting problems with many related time series, as is the focus of the present work, it can be safely assumed that neural network are the state of the art. For example, the models described in~\cite{smyl2018m4} won the recent M4 forecasting competition~\cite{makridakis2018m4} by a large margin. Most recent work on forecasting using neural networks focusses primarily on novel or extended network architectures.



Input transformations have a long and rich history in time series, potentially starting 
with~\citet{Box.Cox1964} who propose a power-transformation of the data to make it "more normal". However, the use of some of the input transformations in the focus of this paper, such as input scaling or variants of binning, are partially folkloric (i.e.\ commonly used in practice by machine learning practitioners but seldomly thoroughly described and investigated).
In contrast, the more general area of probability integral transformation and copula approaches (e.g.,~\cite{copula,copula_review}) enjoys
continued attention. For example,~\citep{Sal2019} propose a semi-parametric neural forecasting model that uses the marginal empirical CDFs combined with Gaussian copulas to model non-Gaussian multivariate data.
In order to be tractable, it assumes a particular low-rank structure of the covariance structure.

Another set of approaches for modeling the output distributions, related to using a categorical distribution on binned time series values, are techniques based on quantile regression \citep{koenker1978regression, koenker2005quantile,wen2017}. In these approaches, instead of modeling the entire output distribution, only a fixed set of quantile levels is predicted. The spline quantile function approach of \citet{gasthaus2019probabilistic} that we compare in our study is an extension of these techniques, where the quantile levels to be predicted a learned by the model and interpolated using a linear spline.

The idea of global-local models, i.e.\ models that explicitly model the patterns shared between time series globally (i.e.\ across time series), while allowing the idiosyncratic behavior of each time series to be modeled locally (i.e.\ per time series), have also been explored \citep{wang2019deepfactors, sen2019think, KDD19Streaming}. The data transformations explored here, which locally apply a transformation before modeling the result globally, can be seen as an instance of the same paradigm. Further, the core idea behind the hybrid binning strategy (Section~\ref{sec:method}) is to mix global (to the panel of time series) and local (specific to a member of the panel) effects.

\section{Conclusions and Future Work}
\label{sec:con}
We have conducted a large-scale study comparing the performance of different input and output transformations when combined with several different types of models. Our investigation shines light on the question to which extent such transformations affect the predictive performance of different model architectures, with the overarching conclusion that carefully choosing and tuning the input and output transformations is important, as it has a large impact on the models' predictive performance, potentially larger than the performance difference between model architectures.


The work presented here can be extended in multiple directions: First and foremost, there are interesting additional kinds of input and in particular output transformations that we want to explore, e.g.\ hybrid binnings using multiple scales, and using hybrid binnings also at the output (e.g.\ using a multi-resolution approach similar to the ``dual softmax'' used in \citep{WaveRNN}). On the methodological side, extensive and principled hyperparameter tuning would allow us to make stronger conclusions about the effectiveness of particular model classes when combined with different input/output representations. 


Finally, categorical sequence data is common in other domains, e.g.\ text in NLP or quantized audio data in speech recognition and generation---large sub-fields of AI where novel deep learning techniques are constantly developed and improved.
Modifying models from those domains to fit the forecasting problem better is a productive line of recent research, e.g.,~\cite{fan2019multi,li2019}. Exploring whether models from these domains can perform well in the forecasting setting without substantial modifications by discretizing the inputs using the techniques discussed here is an interesting open question, that---if answered affirmatively---would allow further improvements 
in these domains to immediately carry over to the time series domain. However, it is still surprising to us how well the categorical distribution performs, even though it ignores the order in data, and this needs further understanding. The discretized logistic mixture likelihood \citep{PixelCNN} has been proposed as an alternative to the categorical distribution that retains the ordering. Exploring such methods that can retain the apparent benefits seen with discretized inputs while making use of the order and distance information in the setting of time series forecasting is an interesting avenue for further research.



\end{document}